%% file: med_evo.tex
\documentclass{article}
\usepackage{amsmath}
\usepackage{multirow}
\usepackage{booktabs}
\usepackage{makecell}
\usepackage{array}
\usepackage{caption}
\usepackage{subcaption}
\usepackage{graphicx}
\usepackage{listings}
\usepackage{wrapfig} 
\usepackage[most]{tcolorbox}
\usepackage{pdfcomment}
\usepackage[most]{tcolorbox}
\usepackage{xcolor}
\usepackage[table]{xcolor}
\definecolor{rowhl}{gray}{0.92}

\usepackage[table]{xcolor}   %
\definecolor{rowhl}{gray}{0.92}

\usepackage[most]{tcolorbox}
\definecolor{hazard}{HTML}{B23A48}     %
\definecolor{correct}{HTML}{2D6A4F}    %
\definecolor{boxframe}{gray}{0.55}
\newcommand{\rd}[1]{\textcolor{hazard}{#1}}
\newcommand{\gn}[1]{\textcolor{correct}{#1}}
\usepackage{tabularx}
\usepackage{float}

\PassOptionsToPackage{numbers,compress}{natbib}

 \usepackage[eandd, final]{neurips_2026}

\usepackage[utf8]{inputenc} %
\usepackage[T1]{fontenc}    %
\usepackage{hyperref}       %
\usepackage{url}            %
\usepackage{booktabs}       %
\usepackage{amsfonts}       %
\usepackage{nicefrac}       %
\usepackage{microtype}      %
\usepackage{etoc}           %
\usepackage{placeins}       %
\usepackage{xcolor}         %
\usepackage{algorithm}      %
\usepackage{algpseudocode}  %
\usepackage{pifont}         %
\definecolor{checkgreen}{RGB}{46,125,50}
\definecolor{crossred}{RGB}{198,40,40}

\definecolor{darkgreen}{RGB}{0,180,0}
\hypersetup{
  colorlinks=true,
  linkcolor=magenta,
  citecolor=darkgreen,
  urlcolor=blue
}

\usepackage{algorithm}
\usepackage{algpseudocode}
\usepackage{xcolor}
\usepackage{enumitem}

\definecolor{promptbg}{gray}{0.97}
\definecolor{promptframe}{gray}{0.80}
\definecolor{prompttitlebg}{RGB}{229,239,247}
\definecolor{prompttitlefg}{RGB}{32,64,96}

\lstdefinestyle{promptstyle}{%
  basicstyle=\ttfamily\scriptsize,
  breaklines=true,
  breakatwhitespace=false,
  breakautoindent=false,
  breakindent=0pt,
  columns=fullflexible,
  keepspaces=true,
  showstringspaces=false,
  frame=none,
  backgroundcolor=,          %
  framerule=0.6pt,
  framesep=4pt,
  xleftmargin=0pt,
  xrightmargin=0pt,
  aboveskip=0.2em,
  belowskip=0.2em,
  literate={—}{{---}}3 {–}{{--}}2 {→}{{$\rightarrow$}}1
}

\lstnewenvironment{promptblock}[1][]{\lstset{style=promptstyle,#1}}{}

\definecolor{promptframe}{gray}{0.75}
\definecolor{prompttitlebar}{gray}{0.45}

\newtcblisting{promptbox}[2][]{%
  listing only,
  breakable,
  enhanced,
  colback=white,
  colframe=prompttitlebar,
  colbacktitle=prompttitlebar,
  coltitle=white,
  fonttitle=\bfseries\small,
  title={#2},
  boxrule=0.8pt,
  arc=2pt,
  left=4pt, right=4pt, top=4pt, bottom=4pt,
  before skip=4pt, after skip=4pt,
  listing options={
    style=promptstyle,
    frame=none,
    backgroundcolor=,
    aboveskip=0pt,
    belowskip=0pt,
    #1
  }
}

\newcommand{\realicu}{{\fontsize{0.95em}{1em}\selectfont
{\fontfamily{qpl}\selectfont\textbf{\textls[6]{RealICU}}}}}

\title{\realicu: Do LLM Agents Understand Long-Context ICU Data? A Benchmark Beyond Behavior Imitation}

\author{%
\textbf{Chengzhi Shen}$^{1,2,9,10}$ \quad 
\textbf{Weixiang Shen}$^{1,2,3}$ \quad
\textbf{Tobias Susetzky}$^{1,2}$ \\
\textbf{Chen (Cherise) Chen}$^{4}$ \quad 
\textbf{Jun Li}$^{1,9}$ \quad
\textbf{Yuyuan Liu}$^{5}$ \quad
\textbf{Xuepeng Zhang}$^{6}$ \\
\textbf{Zhenyu Gong}$^{7,\dagger}$ \quad
\textbf{Daniel Rueckert}$^{1,2,8,9,10,\dagger}$ \quad 
\textbf{Jiazhen Pan}$^{1,2,9,\dagger}$ 
\\\\
$^1$Technical University of Munich (TUM)\quad
$^2$TUM University Hospital \quad 
$^3$LMU Munich  \\
$^4$University of Sheffield \quad
$^5$University of Oxford \quad
$^6$Zhongshan Hospital Fudan University \\
$^7$Sun Yat-sen University Cancer Center \quad
$^8$Imperial College London \\
$^9$Munich Center for Machine Learning (MCML) \\
$^{10}$relAI – Konrad Zuse School of Excellence in Reliable AI \\[0.2em]
$^{\dagger}$Corresponding Authors\\ 
}

\begin{document}

\maketitle

\begin{abstract}

Intensive care units (ICU) generate long, dense and evolving streams of clinical information, where physicians must repeatedly reassess patient states under time pressure, underscoring a clear need for reliable AI decision support. Existing ICU benchmarks typically treat historical clinician actions as ground truth. However, these actions are made under incomplete information and limited temporal context of the underlying patient state, and may therefore be suboptimal, making it difficult to assess the true reasoning capabilities of AI systems. We introduce RealICU, a hindsight-annotated benchmark for evaluating large language models (LLMs) under realistic ICU conditions, where labels are created after senior physicians review the full patient trajectory. We formulate four physician-motivated tasks: assess \emph{Patient Status}, \emph{Acute Problems}, \emph{Recommended Actions}, and \emph{Red Flag} actions that risk unsafe outcomes. We partition each trajectory with 30-min windows and release two datasets: RealICU-Gold with 930-window annotations from 94 MIMIC-IV patients, and RealICU-Scale with 11,862 windows extended by \emph{Oracle}, a physician-validated LLM hindsight labeler. Existing LLMs including memory-augmented ones performed poorly on RealICU, exposing two failure modes: a recall-safety tradeoff for clinical recommendations, and an anchoring bias to early interpretations of the patient. We further introduce ICU-Evo to study structured-memory agents that improves long-horizon reasoning but does not fully eliminate safety failures. Together, RealICU provides a clinically grounded testbed for measuring and improving AI sequential decision-support in high-stakes care. Project page: \href{https://chengzhi-leo.github.io/RealICU-Bench/}{chengzhi-leo.github.io/RealICU-Bench}

\end{abstract}

\input{chapters/1_intro}

\input{chapters/2_related_work}

\input{chapters/3_dataset_description}

\input{chapters/4_method}
\input{chapters/5_evaluation}

\input{chapters/6_conclusion}

\clearpage

\bibliographystyle{plainnat}
\bibliography{ref}

\clearpage
\appendix
\input{chapters/appendix}

\end{document}

%% file: chapters/1_intro.tex
\section{Introduction}

\begin{figure}[t]
\centering
\includegraphics[width=\linewidth]{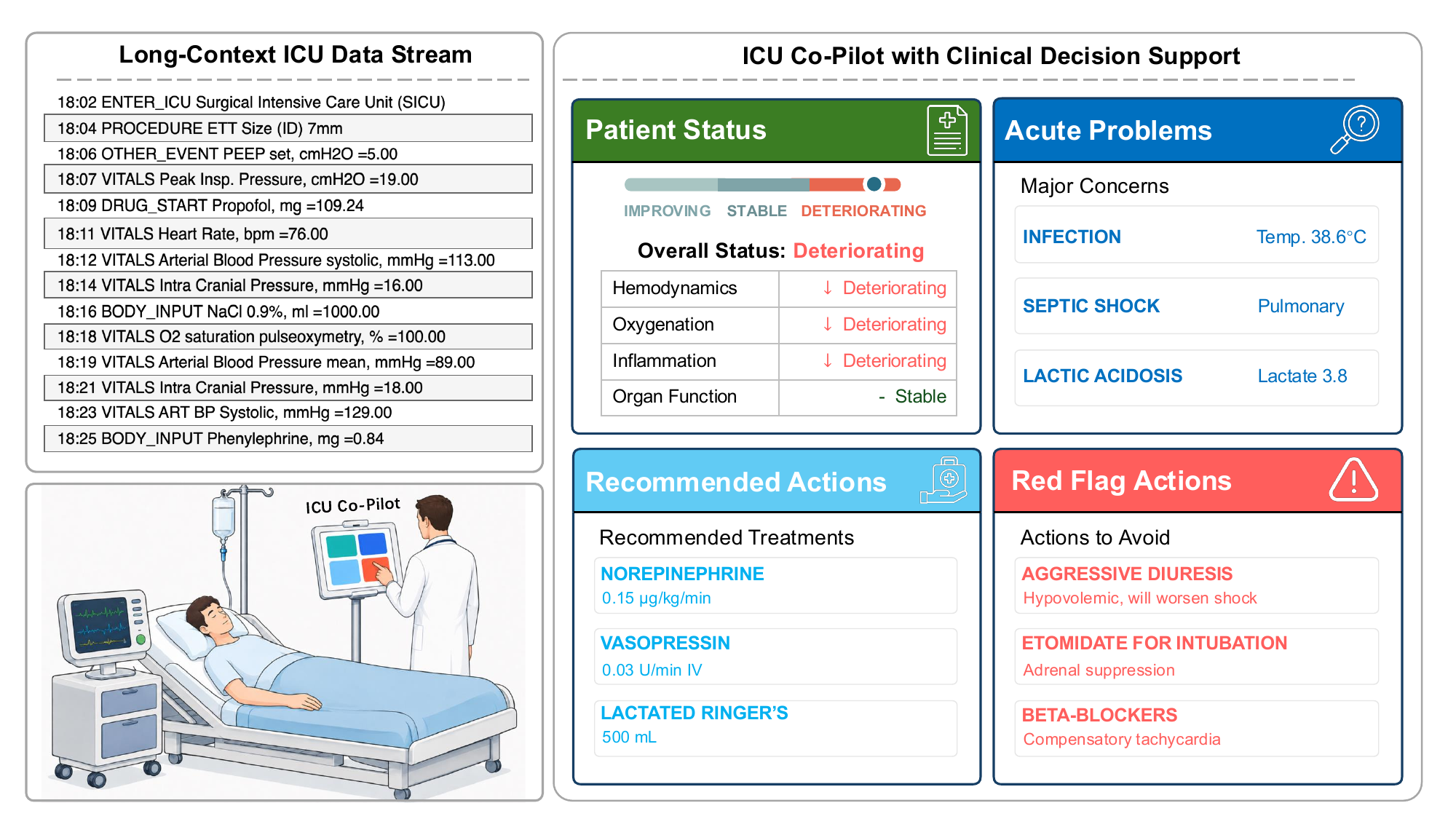}
\caption{ICU decisions are made under massive data volume and time pressure. An ICU AI co-pilot integrates data streams into a decision-support panel that assesses \emph{Patient Status}, identifies \emph{Acute Problems}, proposes \emph{Recommended Actions}, and warns against unsafe \emph{Red Flag} actions.}
\label{fig:main-story}
\end{figure}

The Intensive Care Unit (ICU) is one of the most information-dense environments in the hospital. Within hours, a single patient can generate large volumes of laboratory results, vital signs, medications, nursing observations, and imaging reports~\cite{manor2008quantifying,pickering2010novel}. Physicians must integrate this evolving stream under time pressure, where each measurement captures only a partial slice of the patient's physiological state, and decisions made in one moment may shape outcomes hours or days later~\cite{paul2023effect,rosa2019effects}. This underscores a clear need for AI decision support system in real-time monitoring and decision-making in the ICU, which usually acts as a clinical co-pilot. In consultations with over 30 board-certified clinicians, including five senior ICU physicians who later served as annotators, four capabilities emerged as core requirements for a useful ICU co-pilot: assess \emph{Patient Status}, identify \emph{Acute Problems}, propose \emph{Recommended Actions}, and warn against \emph{Red Flag} actions that may cause unsafe outcomes. Figure~\ref{fig:main-story} illustrates the use case of an AI co-pilot in ICU decision support.

\textbf{Benchmark gap.} Despite rapid progress in Large Language Models (LLMs) and agentic systems, few benchmarks evaluate these four capabilities in real-world ICU settings. Most clinical benchmarks reduce clinical reasoning to static question answering, diagnosis, or summarization~\cite{ma2024clibench, van2023yet, jin2021disease, jin2019pubmedqa, chiu2025simulating}, or to single-endpoint prediction (e.g., mortality~\cite{zhao2020prediction}, shock~\cite{ghosh2017septic,yee2019data}, or acute kidney injury~\cite{malhotra2017risk,dong2021machine}). Such benchmarks aggregate clinical care into isolated predictions, offering little signal on whether a model can reason across a changing patient trajectory. More importantly, benchmarks built on electronic health record (EHR) databases such as MIMIC-IV~\cite{johnson2023mimic}, HiRID~\cite{hyland2020early}, and eICU-CRD~\cite{pollard2018eicu} treat recorded clinician actions as ground-truth labels. But this assumption is fragile. A recorded action reflects what clinicians believed best given incomplete information at the bedside, whereas the optimal action often becomes clear only after reviewing the trajectory using hindsight. Evaluating AI models against such labels therefore rewards behavioral imitation rather than clinical correctness.

\textbf{Proposed benchmark.} To address this gap, we introduce \emph{RealICU}, a hindsight-grounded benchmark built from MIMIC-IV~\cite{johnson2023mimic} for evaluating LLM-based clinical decision support in the ICU. \emph{RealICU} evaluates four physician-motivated tasks over dense 30-minute windows across the ICU trajectory: \emph{Patient Status}, \emph{Acute Problems}, \emph{Recommended Actions}, and \emph{Red Flags}. At each window, the agent observes only information available up to that time, while labels are produced by hindsight physician judgment over the full trajectory. This design scores agents on clinical correctness rather than on recorded behavior. \emph{RealICU} contains two subsets. \emph{RealICU-Gold} provides 930 physician-labeled windows from 94 ICU stays, and \emph{RealICU-Scale} extends evaluation to 11{,}862 windows using \emph{Oracle}, a physician-validated LLM-based hindsight evaluator calibrated against expert consensus.

\textbf{Failure mode identification and mitigation.} Using \emph{RealICU}, we benchmark frontier LLM-based ICU agents across diverse context configurations including memory. Current agents show poor reliability over long ICU contexts, with two failure modes: (i) Recall-safety tradeoff, where higher recommendation recall comes with up to 47.3\% of these recommendations flagged as potentially harmful; (ii) Anchoring bias, where agents preserve early interpretations of the patient despite later contradictory evidence. To mitigate these, we introduce ICU-Evo, a structured-memory agent framework that maintains recent observations, temporal trends, critical events, trajectory summaries, and patient-specific insights. ICU-Evo is backbone-agnostic and improves clinical reasoning, but its safety failures show that structured memory alone is insufficient for reliable ICU co-pilots.

Our key contributions are as follows:

\begin{itemize}[leftmargin=2.0em]

    \item We formulate ICU co-pilot evaluation around four physician-motivated tasks: \emph{Patient Status}, \emph{Acute Problems}, \emph{Recommended Actions}, and \emph{Red Flags}. Unlike static clinical QA or outcome prediction benchmarks, these tasks evaluate whether an AI system can support continuous bedside reassessment across an evolving ICU trajectory.

    \item We release \emph{RealICU}, a hindsight-annotated benchmark for clinical correctness rather than behavioral imitation. Agents observe only data available at decision time, while labels are produced by hindsight physician judgment over the full trajectory. \emph{RealICU-Gold} provides 930 physician-consensus windows from 94 ICU stays, and \emph{RealICU-Scale} extends this to 11{,}862 windows using \emph{Oracle}, a physician-validated LLM-based hindsight evaluator.

    \item We identify gaps in current LLM ICU agents and study structured memory as a mitigation. Across frontier LLMs and multiple context strategies, \emph{RealICU} remains largely unsolved. We identify a recall--safety tradeoff and anchoring bias as major failure modes, and introduce ICU-Evo, a structured-memory agent that improves long-horizon reasoning but shows that memory alone is insufficient for safe ICU decision support.

\end{itemize}

%% file: chapters/2_related_work.tex
\section{Related Work}

\paragraph{Clinical Benchmarks for LLMs and Agents.} Exam-style benchmarks such as MedQA~\cite{jin2021disease}, PubMedQA~\cite{jin2019pubmedqa}, and MedXpertQA~\cite{zuo2025medxpertqa} evaluate clinical knowledge as multiple-choice recall under complete information, a format well-addressed by state-of-the-art models that reveals little about decisions under uncertainty. Conversational benchmarks such as AI Hospital~\cite{fan2025ai}, AgentClinic~\cite{schmidgall2024agentclinic}, and VivaBench~\cite{chiu2025simulating} require agents to gather history, order investigations, and converge on a diagnosis over multiple turns, exposing failure modes such as premature diagnostic closure. MedAgentBench~\cite{jiang2025medagentbench} moves closer to real EHR environments but retains a task-completion framing rather than evaluating overall patient management. None of these benchmarks evaluates sequential decision-making over long ICU trajectories or distinguishes behavioral imitation from clinical correctness. \emph{RealICU} addresses both by grounding evaluation in hindsight physician judgment over the full ICU trajectory, providing dense and trajectory-level signal of clinical correctness.

\paragraph{Memory-Augmented LLM Agents.} Recent LLM agent architectures have explored a range of memory designs. ReAct~\cite{yao2022react} appends all reason-action results sequentially but saturates quickly as context accumulates. AgentFold~\cite{ye2025agentfold} addresses this by summarizing completed sub-tasks at multiple temporal scales. Evo-Memory~\cite{wei2025evo} unifies reasoning, action, and memory refinement in a test-time loop. Retrieval-based systems such as RAG~\cite{arslan2024survey,cuconasu2024power} and A-MEM~\cite{xu2025mem} enable selective access over long histories. However, these systems treat clinical context equally, making no distinction between static patient background~\cite{mattey2022hospitalised}, time-sensitive physiological trends~\cite{li2014physiological}, and high-level trajectory~\cite{sousa2020developmental,reed2015defining}, which play fundamentally different roles in clinical reasoning. ICU-Evo organizes clinical context into heterogeneous memory types aligned with these distinctions, enabling systematic study of how structured memory design shapes ICU decision-making.

%% file: chapters/3_dataset_description.tex
\section{\emph{RealICU} Benchmark}

\begin{figure}[t]
\centering
\includegraphics[width=\linewidth]{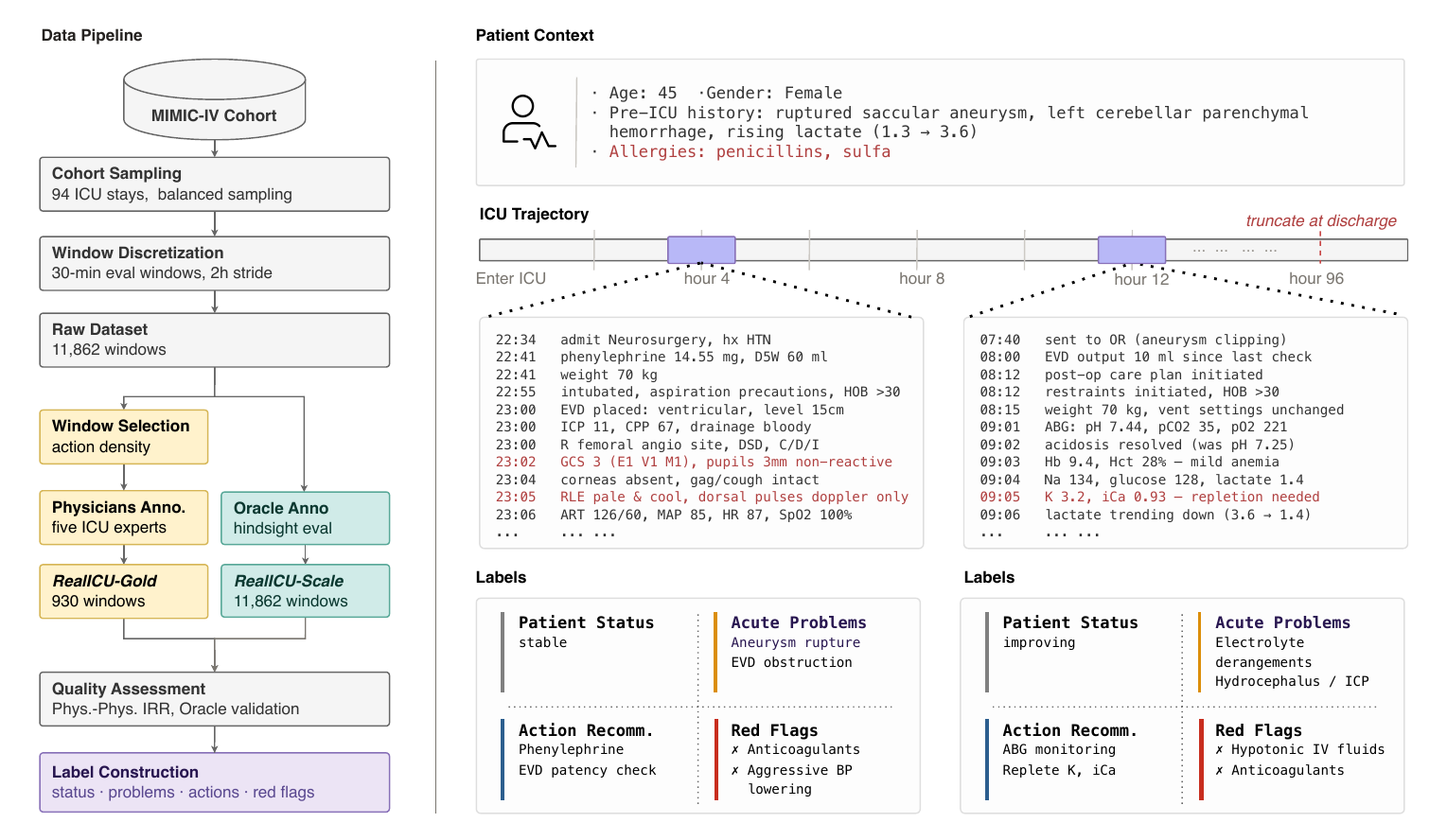}
\caption{Left: Data pipeline for \emph{RealICU-Gold} and \emph{RealICU-Scale}. Right: Data samples for a patient ICU trajectory. For each evaluation window, \emph{RealICU} provides raw observation data and clinical labels, including patient status, acute problems, action recommendation, and red flag action.}
\label{fig:data-pipeline}
\end{figure}

\emph{RealICU} evaluates LLM agents on sequential clinical decision-making across ICU trajectories, mirroring standard medical quality review: model outputs are assessed against hindsight physician labels produced with full knowledge of patient trajectory rather than against logged clinician actions.

\emph{RealICU} consists of two datasets. \emph{RealICU-Gold} contains 930 sparsely sampled windows from 94 ICU stays labeled by physician consensus. To scale beyond manual annotation, we introduce \emph{Oracle}, an LLM-based hindsight evaluator validated against \emph{RealICU-Gold}, yielding \emph{RealICU-Scale} with 11{,}862 densely labeled windows. Both datasets are released test-only to prevent leakage. Detailed statistics are in Figure~\ref{fig:cohort_stats}, Figure~\ref{fig:realicu_gold_stats}, and Figure~\ref{fig:realicu_scale_stats}.

Each window $W_t = (X_t;\; S_t, P_t, A_t, R_t)$ contains clinical observations up to time $t$, annotated for four tasks: \emph{Patient Status} $S_t$, \emph{Acute Problems} $P_t$, \emph{Recommended Actions} $A_t$, and \emph{Red Flag Actions} $R_t$. The model predicts $(\hat{S}_t, \hat{P}_t, \hat{A}_t)$ from $X_t$; $R_t$ serves as a safety check against $\hat{A}_t$. This asymmetry between partial observation and hindsight annotation mirrors the gap between real-time decision-making and hindsigth review. Figure~\ref{fig:data-pipeline} illustrates the data construction pipeline and samples.

\subsection{Dataset Construction}

\paragraph{Cohort.} We sample 94 ICU stays from the MIMIC-IV~\cite{johnson2023mimic} cohort, each from a distinct patient and balanced by ICU outcome. Stays shorter than 4 hours are discarded. To capture both early stabilization and long trajectories, we balance stays by duration above and below 96 hours. 

\emph{RealICU} includes timestamped vital signs, laboratory results, medications, procedures, diagnoses, clinical notes, and radiology reports. Raw imaging and other multi-modal data are excluded, while their metadata and text reports are retained when available. See Appendix~\ref{sec:app_dataset_details} for preprocessing details.

\paragraph{Windowing.} We define 30-minute windows to capture short-term clinical dynamics. This resolution is motivated by the observed frequency of clinical actions: the average interval between consecutive actions is 22.3 minutes, and 80.3\% of subsequent actions occur within 30 minutes. We sample evaluation checkpoints every 2 hours to reduce redundancy while retaining all intermediate observations. Sensitivity analysis with 30-minute and 1-hour strides shows stable performance, with at most 0.017 absolute variation across primary metrics.

\subsection{Tasks}
\label{sec:tasks}
We identify four crucial ICU reasoning tasks below after consulting more than 30 clinicians, including five senior ICU physicians who later served as annotators. Together they cover the key capabilities of a useful ICU co-pilot. For all four tasks, each prediction is accompanied by supporting evidence $\mathcal{E} \subseteq X_t$ drawn from the raw events in the recorded history.

\noindent\textbf{\emph{Patient Status}.} A classification of whether the patient is improving, stable, or deteriorating relative to recent context: $S_t=(s_t,\mathcal{E}_t)$, where $s_t \in \{\texttt{improving}, \texttt{stable}, \texttt{deteriorating}\}$ and $\mathcal{E}_t \subseteq X_t$.

\noindent\textbf{\emph{Acute Problems}.} A free-text set of acute problems or emerging risks that require active management: $P_t=\{(p_i,\mathcal{E}_i)\}_{i=1}^{k}$, where $\mathcal{E}_i \subseteq X_t$.

\noindent\textbf{\emph{Action Recommendation}.} A free-text set of actions likely to benefit the patient within one hour, such as stabilizing physiology or preventing deterioration: $A_t=\{(a_j,\mathcal{E}_j)\}_{j=1}^{m}$, where $\mathcal{E}_j \subseteq X_t$. The one-hour horizon focuses on immediate clinical interventions rather than longer-term care planning, following recommendations from ICU physicians.

\noindent\textbf{\emph{Red Flags}.} A free-text set of high-risk actions that should be avoided because they may be harmful under the patient's current physiology or trajectory: $R_t=\{(r_l,\mathcal{E}_l)\}_{l=1}^{n}$, where $\mathcal{E}_l \subseteq X_t$.

\subsection{Annotation Protocol}
\label{sec:pipeline}

\paragraph{\emph{RealICU-Gold} with physician consensus.}

We begin from sampling approximately 10 windows per ICU stay by action density $\rho_t = |\mathcal{E}^{\text{action}}_t| / |\mathcal{E}_t|$, i.e. the fraction of action events inside each window. We draw 80\% of windows from the $\rho_t \geq 0.5$ regime, where interventions are frequent, and 20\% from $\rho_t < 0.5$ as a control set. Each window is independently labeled by at least two of five senior ICU physicians. Inter-rater reliability (IRR) among physicians ranges from 0.826 to 0.985 across the four tasks (Table~\ref{tab:gold-oracle}), confirming both strong label reproducibility and that the task definitions are sufficiently precise for consistent clinical judgment.  Windows without physician agreement are dropped, yielding 930 validated windows in \emph{RealICU-Gold}.

\begin{wraptable}{r}{0.4\linewidth}
\vspace{-\baselineskip}
\setlength{\tabcolsep}{4pt}
\renewcommand{\arraystretch}{1.5}
\small
\begin{minipage}{\linewidth}
\caption{\emph{RealICU-Gold} label quality and \emph{Oracle} validation.}
\label{tab:gold-oracle}
\centering
\begin{tabular}{lcc}
\toprule
\textbf{Task} & \textbf{Phys. IRR} & \textbf{Oracle F1} \\
\midrule
\emph{Patient Status}  & 0.985 & 0.987 \\
\emph{Acute Problems} & 0.980 & 0.987 \\
\emph{Action Recom.}   & 0.826 & 0.895 \\
\emph{Red Flags}       & 0.916 & 0.964 \\
\bottomrule
\end{tabular}
\end{minipage}
\end{wraptable}

\paragraph{\emph{RealICU-Scale} with \emph{Oracle} scaling.}
Despite high quality, manual annotation covers only a sparse sample of each ICU stay. We therefore introduce \emph{Oracle}, an LLM evaluator operating under the same hindsight conditions as the physicians, and apply it to densely label every window across the cohort, yielding 11{,}862 annotated windows in \emph{RealICU-Scale}. We validate \emph{Oracle} by measuring its F1 score against physician consensus on \emph{RealICU-Gold}. \emph{Oracle} achieves at least 0.895 F1 score across all four tasks (Table~\ref{tab:gold-oracle}), supporting its use as a reliable hindsight annotator at scale. While \emph{Oracle} is backbone-agnostic, we instantiate it with Gemini-3.1-pro~\cite{Gemini31Pro2026} in this work. Detailed \emph{Oracle} prompt is in Appendix~\ref{app:prompts}.

\paragraph{Label construction.}
Labels for \emph{Patient Status}, \emph{Acute Problems}, and \emph{Red Flags} are taken directly from annotations. For \emph{Action Recommendation}, we restrict the annotation space to critical clinical interventions, discarding routine monitoring. Annotators review each action as \texttt{best-practice}, \texttt{acceptable}, or \texttt{potentially-harmful}, and may add free-text actions that should have been taken but were not observed. $A_t$ is constructed as the union of \texttt{best-practice} and \texttt{acceptable} actions together with these free-text additions. \emph{Red Flags} are annotated independently as a separate label, not derived from \texttt{potentially-harmful} actions.

Among 2,196 clinically informative recorded actions, physicians rated 72.45\% as best practice, 22.36\% as acceptable, and 5.19\% as potentially harmful. At the window level, 93 of 930 windows (10.0\%) contained at least one potentially harmful recorded action. These findings highlight the limitations of treating historical clinician actions as ground-truth recommendations. Detailed Dataset statistics are shown in Appendix~\ref{sec:app_dataset_statistics}.

\subsection{Evaluation Framework}
\label{sec:eval}
A model under test $\mathcal{M}$ maps observations $X_t$ to predictions $(\hat{S}_t, \hat{P}_t^{(k)}, \hat{A}_t^{(k)})$, where $\hat{P}_t^{(k)}$ and $\hat{A}_t^{(k)}$ are top-$k$ ranked lists, with access only to events up to time $t$. In this paper we focus on LLM agents, but $\mathcal{M}$ can be any model. Models are evaluated against \emph{RealICU-Gold} and \emph{RealICU-Scale}, providing sparse gold-standard supervision and trajectory-level evaluation at scale respectively. Algorithm~\ref{alg:eval} summarizes the complete evaluation framework.

\paragraph{Semantic matching.}
To score free-text tasks (\emph{Acute Problems}, \emph{Recommended Actions}, \emph{Red Flag Actions}), we adopt PubMedBERT~\cite{gu2021domain} and define a binary match, where $\tau$ is calibrated against 100 expert-annotated pairs, achieving $0.96$ F1 at $\tau=0.5$ (Appendix~\ref{app:semantic_match}):
\begin{equation}
\mathrm{match}(x_{\text{pred}},\ x_{\text{ref}}) = \mathbf{1}\!\left[\cos(\mathbf{e}_{\text{pred}},\, \mathbf{e}_{\text{ref}}) \geq \tau\right].
\end{equation}

\paragraph{Metrics.}
\emph{Patient Status} is evaluated with accuracy and macro-F1 to avoid dominance by the majority class (\texttt{stable}). \emph{Acute Problems} and \emph{Recommended Actions} are set-matching tasks evaluated with Hit@$k$ and Recall@$k$ at $k{=}5$. \emph{Red Flag Actions} serves as a safety check via the Harmful Recommendation Rate (HRR). Let $\mathcal{S}$ be the set of ICU stays, $\mathcal{W}_s$ the windows in stay $s$, $\hat{A}_t^{(k)}$ the top-$k$ recommendations, and $R_t$ the red-flag set at window $t$; HRR averages the fraction of recommended actions that are flagged across stays:
\begin{equation}
\mathrm{HRR}(\mathcal{M}) \;=\;
\frac{1}{|\mathcal{S}|}\sum_{s \in \mathcal{S}}
\frac{\sum_{t \in \mathcal{W}_s} \big|\hat{A}_t^{(k)} \cap R_t\big|}
     {\sum_{t \in \mathcal{W}_s} \big|\hat{A}_t^{(k)}\big|}.
\end{equation}

\definecolor{commentgreen}{rgb}{0.18,0.45,0.45}
\algrenewcommand\algorithmiccomment[1]{%
  \hfill{\color{commentgreen}\(\triangleright\)~#1}%
}

\begin{algorithm}[t]
\caption{\emph{RealICU} Evaluation Framework.}
\label{alg:eval}
\small
\begin{algorithmic}[1]
\Require model $\mathcal{M}$; label source $\mathcal{R} \in \{\emph{Gold},\,\emph{Scale}\}$; ICU stay set $\mathcal{S}$; per-stay window sets $\{\mathcal{W}_s\}_{s\in\mathcal{S}}$
\For{each ICU stay $s \in \mathcal{S}$}
  \State $h \gets 0,\quad n \gets 0$ \Comment{red-flag hits / total recommendations}
  \For{each window $t \in \mathcal{W}_s$ in chronological order}
    \State $(\hat{S}_t,\, \hat{P}_t^{(k)},\, \hat{A}_t^{(k)}) \gets \mathcal{M}(X_t)$ \Comment{model sees events up to $t$}
    \State $(S_t,\, P_t,\, A_t,\, R_t) \gets \mathcal{R}(t)$ \Comment{pre-labeled by hindsight annotator}
    \State evaluate $(\hat{S}_t, S_t)$ for \emph{Patient Status} accuracy and F1
    \State evaluate $(\hat{P}_t^{(k)}, P_t)$ for \emph{Acute Problems} Hit@$k$ / Recall@$k$ \Comment{semantic matching}
    \State evaluate $(\hat{A}_t^{(k)}, A_t)$ for \emph{Action Recommendation} Hit@$k$ / Recall@$k$ \Comment{semantic matching}
    \State $h \mathrel{+}= |\hat{A}_t^{(k)} \cap R_t|;\quad n \mathrel{+}= |\hat{A}_t^{(k)}|$ \Comment{safe recommendation check}
  \EndFor
  \State aggregate per-window scores
\EndFor
\State \Return scores across $\mathcal{S}$ for each task
\end{algorithmic}
\end{algorithm}

%% file: chapters/4_method.tex
\section{ICU-Evo: An ICU Agent System with Evolving Memory}
\label{sec:method}

ICU decision-making is sequential, where the underlying patient state is only partially observable with clinical measurements and can only be updated via new observations. We model this as a partially observable Markov process~\cite{cassandra1998exact,spaan2012partially} and approximate the latent patient state with a structured memory $M_t$. We introduce ICU-Evo as an instance of the memory-augmented agent frameworks to study how structured memory design shapes clinical decision-making.

\subsection{Memory as a Structured Belief State}

Given the context $X_t$ and static patient context $c$ (e.g.\ demographics, allergies, pre-ICU history), ICU-Evo maintains a structured memory state $M_t$ updated at each window by incorporating the new measurements $x_t = X_t - X_{t-1}$, and produces task-specific predictions $y_t^{(k)}$ via
\begin{equation}
M_t = \mathcal{U}(M_{t-1},\, x_t), \qquad y_t^{(k)} = f^{(k)}(M_t, c).
\label{eq:memory}
\end{equation}

The memory decomposes into five components following clinical reasoning:
\begin{equation}
\label{eq:memory_components}
M_t = \bigl\{
M_t^{\mathrm{work}},\;
M_t^{\mathrm{trend}},\;
M_t^{\mathrm{event}},\;
M_t^{\mathrm{traj}},\;
M_t^{\mathrm{insight}}
\bigr\}.
\end{equation}
\emph{Working memory} $M_t^{\mathrm{work}}$ holds the most recent raw observations at detailed resolution. \emph{Trend memory} $M_t^{\mathrm{trend}}$ captures signal trends of vital and lab values. \emph{Critical-event memory} $M_t^{\mathrm{event}}$ is a persistent, append-only log of clinically critical events that change the patient story, such as abnormal physiology, interventions, and turning points. \emph{Trajectory memory} $M_t^{\mathrm{traj}}$ provides a compressed narrative of the stay at periodic intervals. \emph{Insight memory} $M_t^{\mathrm{insight}}$ maintains patient-specific hypotheses constructed as deviations from population-level expectation. Every memory component carries evidence from raw observations, so any clinical decision is explainable and verifiable against the patient record. In Table~\ref{tab:memory}, we summarize the memory components with corresponding agent sources. 

\subsection{ICU-Evo Agent Pipeline}

ICU-Evo realizes the memory update operator $\mathcal{U}$ through three specialized agents operating at different temporal scales over the shared memory. ICU-Evo belongs to a broader family of memory-augmented agent systems. We discuss it alongside recent agent systems in Appendix~\ref{app:icu-evo}. Detailed prompts are reported in Appendix~\ref{app:prompts}.

\paragraph{Observation Agent ($\mathcal{A}_{\text{obs}}$).}
A rule-based agent that turns raw measurements into structured signals at every window. It normalizes units, aligns observations to the 30-minute window grid, and extracts trend signals from vitals using Piecewise Aggregate Approximation~\cite{guo2010improved}:
\begin{equation}
\bigl(M_t^{\mathrm{work}},\, M_t^{\mathrm{trend}}\bigr) = \mathcal{A}_{\text{obs}}(M_{t-1}^{\mathrm{work}},\, M_{t-1}^{\mathrm{trend}},\, x_t).
\end{equation}

\paragraph{Assessment Agent ($\mathcal{A}_{\text{assess}}$).}
For every $k_a$ cumulative windows, an LLM transforms recent observations into a trajectory summary and detects critical events. It consumes the working and trend memory accumulated over the past $k_a$ windows, producing a trajectory summary $z_t$ appended to $M_t^{\mathrm{traj}}$ and critical events $e_t$ appended to $M_t^{\mathrm{event}}$:
\begin{equation}
\bigl(z_t,\, e_t\bigr) = \mathcal{A}_{\text{assess}}\bigl(M_{t-k_a:t}^{\mathrm{work}},\, M_{t-k_a:t}^{\mathrm{trend}}\bigr).
\end{equation}

\paragraph{Insight Agent ($\mathcal{A}_{\text{insight}}$).}
Every $k_i$ windows, an LLM proposes hypotheses about what is driving the patient's clinical course and gathers supporting evidence $e_s$ and counter-evidence $e_r$ from $M_{t}^{\mathrm{event}}$. A hypothesis $h_t$ is accepted if $s(h_t) > r(h_t)$ and rejected otherwise. The Insight Agent actively reasons about patient-specific patterns, such as unusual drug responses or persistent abnormalities, promoting individualized care beyond averaged guidelines: 
\begin{equation}
M_t^{\mathrm{insight}} = \mathcal{A}_{\text{insight}}\bigl(M_{t-1}^{\mathrm{insight}},\, M_{t-k_i:t}^{\mathrm{event}}\bigr).
\end{equation}

\paragraph{Predictor ($f^{(k)}$).}
The predictor is a task-specific prompted LLM over the full memory state and static patient context, decoupled from the agent system (Equation~\ref{eq:memory}).

%% file: chapters/5_evaluation.tex
\section{Evaluation \& Analysis}
\label{sec:evaluation}

\begin{table*}[t]
\centering
\setlength{\tabcolsep}{5pt}
\renewcommand{\arraystretch}{1.18}
\caption{Evaluation results on \emph{RealICU-GOLD}. Within each backbone, \textbf{bold} marks the best system per column and \underline{underline} the second best.}
\label{tab:main-results-gold}
\small
\begin{tabular}{ll cc cc cc c}
\toprule
 & & \multicolumn{2}{c}{\textbf{\textit{Patient Status}}}
   & \multicolumn{2}{c}{\textbf{\textit{Acute Problems}}}
   & \multicolumn{2}{c}{\textbf{\textit{Action Recom.}}}
   & \textbf{\textit{Red Flags}} \\
\cmidrule(lr){3-4} \cmidrule(lr){5-6} \cmidrule(lr){7-8} \cmidrule(lr){9-9}
\textbf{Backbone} & \textbf{System}
  & Acc.$\uparrow$ & F1$\uparrow$
  & Hit@5$\uparrow$ & R@5$\uparrow$
  & Hit@5$\uparrow$ & R@5$\uparrow$
  & HRR@5$\downarrow$ \\
\midrule
\multirow{4}{*}{\textbf{Gemini-3.1-pro~\cite{Gemini31Pro2026}}}
  & Full-context           & 0.298 & \underline{0.258} & 0.486 & 0.308 & 0.259 & 0.152 & \textbf{0.137} \\
  & Local-window           & \underline{0.315} & 0.239 & 0.459 & 0.258 & 0.395 & 0.260 & \underline{0.151} \\
  & RAG                    & 0.402 & 0.348 & \underline{0.596} & \underline{0.342} & \underline{0.496} & \underline{0.313} & 0.216 \\
\rowcolor{rowhl}
  & ICU-Evo       & \textbf{0.459} & \textbf{0.365} & \textbf{0.823} & \textbf{0.526} & \textbf{0.676} & \textbf{0.534} & 0.300 \\
\midrule
\multirow{4}{*}{\textbf{GPT-5.4~\cite{OpenAIGPT54_2026}}}
  & Full-context           & \underline{0.294} & 0.233 & 0.510 & 0.348 & 0.404 & 0.300 & 0.298 \\
  & Local-window           & 0.233 & 0.184 & 0.500 & 0.293 & 0.380 & 0.281 & \textbf{0.165} \\
  & RAG                    & 0.288 & \underline{0.256} & \underline{0.599} & \underline{0.349} & \underline{0.480} & \underline{0.398} & \underline{0.234} \\
\rowcolor{rowhl}
  & ICU-Evo       & \textbf{0.312} & \textbf{0.264} & \textbf{0.867} & \textbf{0.570} & \textbf{0.676} & \textbf{0.534} & 0.473 \\
\midrule
\multirow{4}{*}{\textbf{Qwen3-235B~\cite{yang2025qwen3}}}
  & Full-context           & 0.225 & \underline{0.188} & \underline{0.384} & \underline{0.226} & 0.329 & 0.222 & 0.117 \\
  & Local-window           & 0.152 & 0.154 & 0.213 & 0.126 & 0.352 & 0.242 & \textbf{0.080} \\
  & RAG                    & \textbf{0.315} & \textbf{0.271} & 0.379 & 0.211 & \underline{0.453} & \underline{0.324} & \underline{0.095} \\
\rowcolor{rowhl}
  & ICU-Evo       & \underline{0.253} & 0.197 & \textbf{0.600} & \textbf{0.362} & \textbf{0.526} & \textbf{0.357} & 0.117 \\
\bottomrule
\end{tabular}
\end{table*}

We adopt three LLM backbones: Gemini-3.1-pro~\cite{Gemini31Pro2026}, GPT-5.4~\cite{OpenAIGPT54_2026}, and Qwen3-235B-A22B~\cite{yang2025qwen3}. Gemini-3.1-pro~\cite{Gemini31Pro2026} is also used as the hindsight LLM annotator. We evaluate ICU-Evo on \emph{RealICU-Gold} and \emph{RealICU-Scale} against three baselines sharing the same predictor: (i) \emph{full-context}, all prior observations up to the window; (ii) \emph{local-window}, the current window only; (iii) \emph{RAG}, top-5 windows retrieved via PubMedBERT~\cite{gu2021domain} embeddings. See Appendix~\ref{sec:eval:setup} for experiment details.

\subsection{\emph{RealICU} Remains Unsolved for Current LLM Systems}
\label{sec:eval:realicu}

\emph{RealICU} remains unsolved for current frontier LLMs and agent systems. Across all evaluation setups in Table~\ref{tab:main-results-gold}, ICU-Evo with Gemini-3.1-pro~\cite{Gemini31Pro2026} reaches only $0.459$ accuracy on \emph{Patient Status} and $0.534$ Recall@5 on \emph{Action Recommendation}. More concerning, \emph{Red Flags} HRR@5 stays non-trivial across all configurations, indicating current LLM systems still recommend potentially harmful actions in high-stake ICU setting. Together, these gaps establish \emph{RealICU} as a clinically grounded safety check for future AI decision-support systems.

\begin{figure}[t]
\centering
\includegraphics[width=\linewidth]{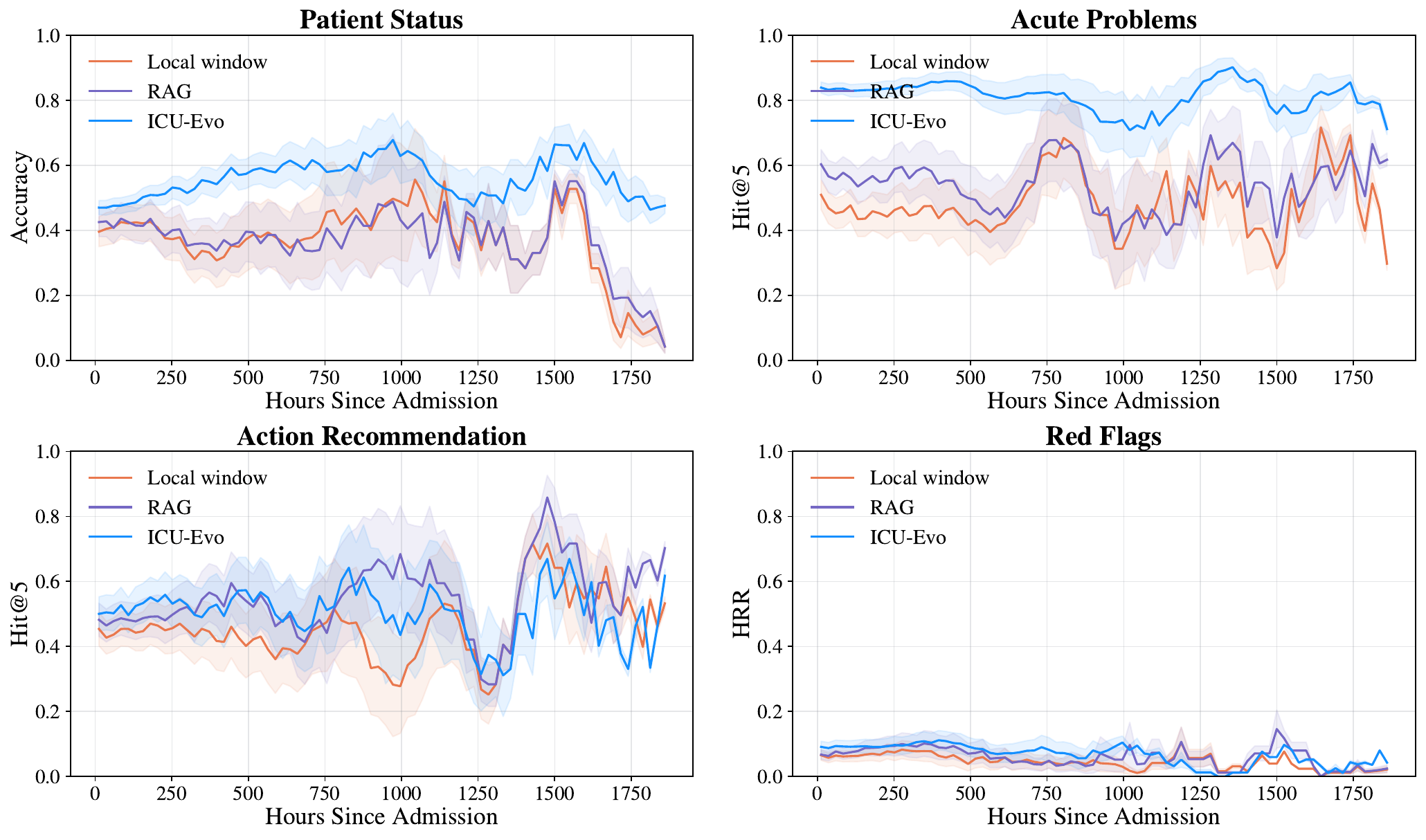}
\caption{Temporal performance on \emph{RealICU-Scale} (Gemini-3.1-pro~\cite{Gemini31Pro2026}). ICU-Evo demonstrates its advantage on \emph{Patient Status} and \emph{Acute Problems} even up to 1,800-hour trajectory. }
\label{fig:realicu-scale}
\end{figure}

\subsection{Structured Memory Improves Clinical Reasoning}
\label{sec:eval:memory}
Structured memory generally improves clinical problem identification and action coverage, although these gains can come at the cost of increased harmful recommendations. With GPT-5.4~\cite{OpenAIGPT54_2026}, ICU-Evo improves over RAG by $26.8$ Hit@5 points on \emph{Acute Problems} and $19.6$ on \emph{Action Recommendation}, with similar margins on Gemini-3.1-pro~\cite{Gemini31Pro2026} and Qwen3-235B~\cite{yang2025qwen3} (Table~\ref{tab:main-results-gold}). The pattern holds on the densely labeled \emph{RealICU-Scale} (Table~\ref{tab:main-results-full}, Figure~\ref{fig:realicu-scale}). ICU-Evo's Hit@5 on \emph{Acute Problems} stays near 0.8 even for stays up to 1,800 hours, while non-memory baselines remain about 20 points lower and visibly noisier. Future ICU decision-support agents will benefit from memory that actively tracks the patient's evolving state and scales to long stays.

\subsection{The Agent-\emph{Oracle} Gap: Beyond Behavioral Imitation}
We observe a large performance gap between Agent and \emph{Oracle} on \emph{RealICU-Gold}. The bottleneck of current ICU agents is not medical knowledge in the LLM backbone but how an agent integrates evidence over time. With Gemini-3.1-pro~\cite{Gemini31Pro2026}, \emph{Oracle} reaches F1 $0.987$ on \emph{Patient Status} and $0.964$ on \emph{Red Flags} identification (Table~\ref{tab:gold-oracle}), while ICU-Evo on the same backbone reaches only $0.365$ F1 on \emph{Patient Status}, with a concerningly high rate of harmful recommendations with $0.300$ HRR (Table~\ref{tab:main-results-gold}). The four clinical tasks are therefore well handled given the full trajectory but break down under real-time conditions. This gap also indicates the value of hindsight evaluation, since scoring agents against recorded clinician actions can only measure how closely the agents imitate human behavior under limited information. Progress on ICU decision support therefore depends on both stronger real-time reasoning architectures and the broader adoption of hindsight evaluation.

\subsection{Ablation Study}
\label{sec:eval:ablation}

We ablate each component of ICU-Evo's memory in a leave-one-out setup (Table~\ref{tab:ablation-memory}). Working memory is crucial for local clinical reasoning, and removing it degrades every task with notable drops on \emph{Acute Problems} Hit@5 (Gemini-3.1-pro~\cite{Gemini31Pro2026}, 0.823 to 0.761). Trajectory memory matters for temporal understanding tasks. Without it, \emph{Acute Problems} and \emph{Action Recommendation} both drop, while \emph{Patient Status} stays stable since it leans on local observations. To further investigate the contribution of structured memory beyond trajectory compression, we compare ICU-Evo with a Trajectory Summary + Local Window baseline, showing additional gains in clinical problem identification and action recommendation (Appendix~\ref{app:trajectory-summary}).

In contrast, insight memory causes fluctuations, and removing it sometimes leads to neutral or beneficial results. This suggests that current LLMs default to medical-generalist priors and is not yet capable to identify reliable personalized clinical patterns across long stays. We examine the failure modes in the next section.

\begin{table}[t]
\centering
\setlength{\tabcolsep}{4pt}
\renewcommand{\arraystretch}{1.15}
\caption{Memory ablation on \emph{RealICU-GOLD}.
Each variant removes one memory component from ICU-Evo, following the notation in Eq.~\ref{eq:memory_components}.
Within each backbone, \textbf{bold} marks the best result and \underline{underline} the second best.}
\label{tab:ablation-memory}
\small
\begin{tabular}{llccccccc}
\toprule
& & \multicolumn{2}{c}{\textbf{\textit{Patient Status}}} & \multicolumn{2}{c}{\textbf{\textit{Acute Problems}}} & \multicolumn{2}{c}{\textbf{\textit{Action Recom.}}} & \textbf{\textit{Red Flags}} \\
\cmidrule(lr){3-4}\cmidrule(lr){5-6}\cmidrule(lr){7-8}\cmidrule(lr){9-9}
\textbf{Backbone} & \textbf{Memory Variant} &
  Acc.$\uparrow$ & F1$\uparrow$ &
  Hit@5$\uparrow$ & R@5$\uparrow$ &
  Hit@5$\uparrow$ & R@5$\uparrow$ &
  HRR$\downarrow$ \\
\midrule
\rowcolor{rowhl}\multirow{6}{*}{\textbf{Gemini-3.1-pro~\cite{Gemini31Pro2026}}}
  & ICU-Evo                     & \underline{0.459} & \textbf{0.365} & \textbf{0.823} & \underline{0.526} & \textbf{0.676} & \underline{0.534} & 0.300 \\
  & \quad w/o $M_t^{\mathrm{work}}$    & 0.383 & 0.294 & 0.761 & 0.461 & 0.507 & 0.308 & \textbf{0.087} \\
  & \quad w/o $M_t^{\mathrm{trend}}$             & 0.451 & 0.352 & 0.811 & 0.527 & 0.521 & 0.330 & 0.097 \\
  & \quad w/o $M_t^{\mathrm{event}}$   & 0.445 & 0.351 & 0.819 & 0.527 & 0.528 & 0.328 & 0.099 \\
  & \quad w/o $M_t^{\mathrm{traj}}$        & 0.443 & 0.362 & 0.789 & 0.500 & 0.506 & 0.304 & 0.090 \\
  & \quad w/o $M_t^{\mathrm{insight}}$           & \textbf{0.462} & 0.356 & \textbf{0.823} & \textbf{0.534} & 0.555 & \underline{0.357} & \underline{0.088} \\
\midrule
\rowcolor{rowhl}\multirow{6}{*}{\textbf{Qwen3-235B~\cite{yang2025qwen3}}}
  & ICU-Evo                     & 0.253 & 0.197 & \textbf{0.600} & \textbf{0.362} & 0.526 & 0.357 & \textbf{0.117} \\
  & \quad w/o $M_t^{\mathrm{work}}$    & 0.159 & 0.127 & 0.421 & 0.233 & 0.447 & 0.307 & \underline{0.117} \\
  & \quad w/o $M_t^{\mathrm{trend}}$             & 0.248 & 0.188 & 0.552 & 0.333 & 0.559 & 0.393 & 0.122 \\
  & \quad w/o $M_t^{\mathrm{event}}$   & 0.236 & 0.187 & 0.546 & 0.320 & \underline{0.595} & \underline{0.420} & 0.128 \\
  & \quad w/o $M_t^{\mathrm{traj}}$        & \textbf{0.270} & \textbf{0.249} & 0.486 & 0.290 & 0.557 & 0.393 & 0.138 \\
  & \quad w/o $M_t^{\mathrm{insight}}$           & \underline{0.250} & \underline{0.202} & \underline{0.587} & \underline{0.348} & \textbf{0.601} & \textbf{0.420} & 0.141 \\
\bottomrule
\end{tabular}
\end{table}

\subsection{Failure Mode Analysis}
\label{sec:eval:failure}

\paragraph{\emph{Oracle} failure modes.} \emph{Oracle} reaches around $90\%$ F1 across all tasks (Table~\ref{tab:gold-oracle}). The remaining disagreements concentrate on two patterns: (i) boundary mis-calibration on \emph{Patient Status}, where failures fall on \texttt{stable}--\texttt{improving} or \texttt{stable}--\texttt{deteriorating} borders; (ii) granularity mismatch on \emph{Acute Problems}, where \emph{Oracle} reaches for broad descriptors (e.g.\ hemodynamic instability) while physicians name specific complications (e.g.\ ventilator-associated pneumonia). These are edge cases rather than systematic errors, supporting \emph{Oracle}'s reliability as a large-scale annotator.

\paragraph{Agent failure modes.}
The most consequential failure is the recall--safety tradeoff, where higher recommendation recall increases the incidence of harmful clinical suggestions. In Table~\ref{tab:main-results-gold}, ICU-Evo (GPT-5.4~\cite{OpenAIGPT54_2026}) gains more than $20$ \emph{Action Recommendation} Hit@5 points over RAG, but its HRR@5 doubles from 0.234 to 0.473. We use an LLM-based classifier to group these 394 cases, and the majority concentrate in four high-stakes families: hemodynamic and pressor management ($n{=}135$), volume and diuresis ($n{=}64$), anti-coagulation ($n{=}54$), and ventilation/sedation ($n{=}53$). We find that current LLM agents tend to recognize part of a syndrome and propose the full treatment bundle before contraindications are ruled out.

The second failure is anchoring bias, where agents over-commit to early interpretations and ignore later evidence. Removing insight memory improves \emph{Action Recommendation} Hit@5 from 0.526 to 0.601 on Qwen3-235B~\cite{yang2025qwen3} (Table~\ref{tab:ablation-memory}), indicating that generated insights actively mislead the agent. The agent maintains around $6$ hypotheses per patient, $80\%$ containing anticipatory exceptions (e.g.\ below-average tolerance, paradoxical response). These priors push the agent toward rescue bundles even when current evidence is weak. Case studies are in Appendix~\ref{app:casestudy}.

%% file: chapters/6_conclusion.tex
\section{Discussion}

\emph{RealICU} reveals a substantial gap between the medical knowledge of frontier LLMs and their ability to reason under partial observability across an evolving ICU trajectory. We identify two recurring failure modes that persist across multiple context configurations. (i) the recall--safety tradeoff, where gains in \emph{Recommended Actions} coverage are accompanied by a higher rate of unsafety. (ii) anchoring bias, where agents commit to an early read of the patient and fail to update as new evidence accumulates. ICU-Evo uses structured, evidence-grounded memory at multiple temporal scales to track the evolving patient state, but multi-scale memory alone does not prevent unsafe recommendations. Reliable ICU co-pilots will require advances in long-context clinical reasoning together with better safety mechanisms.

Beyond the ICU, \emph{RealICU} offers a methodology for evaluating AI systems where recorded human actions are imperfect and the right action is visible only in hindsight. We hope this framing supports broader work on evaluating AI systems in high-stakes sequential decision environments.

\paragraph{Limitations.}
\emph{RealICU} is built on the MIMIC-IV~\cite{johnson2023mimic} cohort, and its demographic and care-pattern distribution may not transfer to ICUs with different staffing or documentation conventions. Extending to multi-center and international data is an important direction. Due to compute constraints, we run a single experiment per LLM configuration and omit variance over long ICU trajectories. We also focus on text-based data, leaving multi-modal data such as imaging and signals to future work. The current benchmark is limited to 94 ICU stays, primarily due to the cost of physician hindsight annotation. 

\section{Acknowledgement}
This paper is supported by the DAAD programme Konrad Zuse Schools of Excellence in Artificial Intelligence, sponsored by the Federal Ministry of Research, Technology and Space. This work is partially funded by the European Research Council (ERC) project Deep4MI (884622).

%% file: chapters/appendix.tex
\appendix

\phantomsection
\section*{Appendix}
\addcontentsline{toc}{section}{Appendix}
\label{sec:appendix}

\subsection*{Appendix Contents}

\begin{itemize}[label={}, leftmargin=0pt]
    \item \hyperref[sec:app_performance_analysis]{\ref*{sec:app_performance_analysis}. \nameref*{sec:app_performance_analysis}}
    \dotfill \pageref{sec:app_performance_analysis}
    \item \hspace{1.5em}\hyperref[sec:app_experiment_setup]{\ref*{sec:app_experiment_setup}. \nameref*{sec:app_experiment_setup}}
    \dotfill \pageref{sec:app_experiment_setup}
    \item \hspace{1.5em}\hyperref[sec:app_realicu_scale_results]{\ref*{sec:app_realicu_scale_results}. \nameref*{sec:app_realicu_scale_results}}
    \dotfill \pageref{sec:app_realicu_scale_results}
    \item \hspace{1.5em}\hyperref[sec:app_averaged_trajectory]{\ref*{sec:app_averaged_trajectory}. \nameref*{sec:app_averaged_trajectory}}
    \dotfill \pageref{sec:app_averaged_trajectory}
    \item \hspace{1.5em}\hyperref[sec:app_per_disease]{\ref*{sec:app_per_disease}. \nameref*{sec:app_per_disease}}
    \dotfill \pageref{sec:app_per_disease}
    \item \hspace{1.5em}\hyperref[app:semantic_match]{\ref*{app:semantic_match}. \nameref*{app:semantic_match}}
    \dotfill \pageref{app:semantic_match}
    \item \hspace{1.5em}\hyperref[app:trajectory-summary]{\ref*{app:trajectory-summary}. \nameref*{app:trajectory-summary}}
    \dotfill \pageref{app:trajectory-summary}
    \item \hspace{1.5em}\hyperref[app:token-efficiency]{\ref*{app:token-efficiency}. \nameref*{app:token-efficiency}}
    \dotfill \pageref{app:token-efficiency}

    \item \hyperref[sec:app_dataset_details]{\ref*{sec:app_dataset_details}. \nameref*{sec:app_dataset_details}}
    \dotfill \pageref{sec:app_dataset_details}
    \item \hspace{1.5em}\hyperref[sec:app_dataset_statistics]{\ref*{sec:app_dataset_statistics}. \nameref*{sec:app_dataset_statistics}}
    \dotfill \pageref{sec:app_dataset_statistics}
    \item \hspace{1.5em}\hyperref[sec:app_cross_validation]{\ref*{sec:app_cross_validation}. \nameref*{sec:app_cross_validation}}
    \dotfill \pageref{sec:app_cross_validation}
    \item \hspace{1.5em}\hyperref[sec:data-preproc]{\ref*{sec:data-preproc}. \nameref*{sec:data-preproc}}
    \dotfill \pageref{sec:data-preproc}

    \item \hyperref[sec:app_icu_evo]{\ref*{sec:app_icu_evo}. \nameref*{sec:app_icu_evo}}
    \dotfill \pageref{sec:app_icu_evo}
    \item \hspace{1.5em}\hyperref[sec:app_icu_evo_formulation]{\ref*{sec:app_icu_evo_formulation}. \nameref*{sec:app_icu_evo_formulation}}
    \dotfill \pageref{sec:app_icu_evo_formulation}
    \item \hspace{1.5em}\hyperref[sec:app_icu_evo_instantiations]{\ref*{sec:app_icu_evo_instantiations}. \nameref*{sec:app_icu_evo_instantiations}}
    \dotfill \pageref{sec:app_icu_evo_instantiations}
    \item \hspace{1.5em}\hyperref[app:icu-evo-system]{\ref*{app:icu-evo-system}. \nameref*{app:icu-evo-system}}
    \dotfill \pageref{app:icu-evo-system}
    \item \hspace{1.5em}\hyperref[sec:app_icu_evo_discussion]{\ref*{sec:app_icu_evo_discussion}. \nameref*{sec:app_icu_evo_discussion}}
    \dotfill \pageref{sec:app_icu_evo_discussion}

    \item \hyperref[sec:app_case_study]{\ref*{sec:app_case_study}. \nameref*{sec:app_case_study}}
    \dotfill \pageref{sec:app_case_study}
    \item \hspace{1.5em}\hyperref[sec:app_recall_safety]{\ref*{sec:app_recall_safety}. \nameref*{sec:app_recall_safety}}
    \dotfill \pageref{sec:app_recall_safety}
    \item \hspace{1.5em}\hyperref[sec:app_anchoring_bias]{\ref*{sec:app_anchoring_bias}. \nameref*{sec:app_anchoring_bias}}
    \dotfill \pageref{sec:app_anchoring_bias}
    \item \hspace{1.5em}\hyperref[sec:memory_snapshot]{\ref*{sec:memory_snapshot}. \nameref*{sec:memory_snapshot}}
    \dotfill \pageref{sec:memory_snapshot}

    \item \hyperref[sec:app_prompts]{\ref*{sec:app_prompts}. \nameref*{sec:app_prompts}}
    \dotfill \pageref{sec:app_prompts}
    \item \hspace{1.5em}\hyperref[sec:app_oracle_prompt]{\ref*{sec:app_oracle_prompt}. \nameref*{sec:app_oracle_prompt}}
    \dotfill \pageref{sec:app_oracle_prompt}
    \item \hspace{1.5em}\hyperref[sec:app_agent_prompt]{\ref*{sec:app_agent_prompt}. \nameref*{sec:app_agent_prompt}}
    \dotfill \pageref{sec:app_agent_prompt}
\end{itemize}

\newpage

\input{chapters/appendix/performance_analysis}
\input{chapters/appendix/dataset_details}
\input{chapters/appendix/icu_evo}

\input{chapters/appendix/case_study}
\input{chapters/appendix/prompts}

%% file: chapters/appendix/performance_analysis.tex
\section{Performance Analysis}
\label{sec:app_performance_analysis}

\subsection{Experiment Setup}
\label{sec:eval:setup}
\label{sec:app_experiment_setup}
We evaluate ICU-Evo on \emph{RealICU-Gold} and \emph{RealICU-Scale} against three baselines sharing the same predictor: (i) \emph{full-context}, all prior observations up to the window; (ii) \emph{local-window}, the current window only; (iii) \emph{RAG}, top-5 windows retrieved via PubMedBERT~\cite{gu2021domain} embeddings. We set $k_a$ and $k_i$ to 12 windows (6 hours) for ICU-Evo. For \emph{Action Recommendation}, we strip the current window's recorded actions before prediction to prevent label leakage. We evaluate every window in \emph{RealICU-Gold} and every fourth window along the trajectory in \emph{RealICU-Scale}.

We use two closed-source LLMs (Gemini-3.1-pro~\cite{Gemini31Pro2026}, GPT-5.4~\cite{OpenAIGPT54_2026}) and one open-source LLM (Qwen3-235B-A22B~\cite{yang2025qwen3}) as backbones for evaluation. Evaluation results on \emph{RealICU-Gold} and \emph{RealICU-Scale} are reported in Table~\ref{tab:main-results-gold} and Table~\ref{tab:main-results-full} respectively. Full-context Gemini and GPT runs on \emph{RealICU-Scale} are omitted due to compute budget on stays beyond hundreds of hours. \emph{Oracle} uses Gemini-3.1-pro~\cite{Gemini31Pro2026} to generated hindsight annotations with access to the full patient trajectory.

\subsection{Evaluation Results on \emph{RealICU-Scale}}
\label{sec:app_realicu_scale_results}

Table~\ref{tab:main-results-full} reports full evaluation results on \emph{RealICU-Scale} across all three backbones and four systems. Full-context evaluation is omitted for Gemini-3.1-pro~\cite{Gemini31Pro2026} and GPT-5.4~\cite{OpenAIGPT54_2026} due to prohibitive inference cost over multi-day ICU trajectories. Qwen3-235B~\cite{yang2025qwen3} is included as a reference open-weight upper bound.

The results on \emph{RealICU-Scale} largely recapitulate the pattern observed on
\emph{RealICU-GOLD} (Table~\ref{tab:main-results-gold}). ICU-Evo achieves the strongest performance on \emph{Acute Problems} and \emph{Action Recommendation} across all three backbones, with particularly large margins on \emph{Acute Problems} Hit (up to $+0.268$ over RAG for Gemini-3.1-pro~\cite{Gemini31Pro2026}). The \emph{Red Flag} HRR remains the consistent weak point of ICU-Evo regardless of backbone, suggesting the same premature anchoring failure mode (see Sec.~\ref{sec:eval:failure}), where current agent systems over-commit to early interpretation of the patient instead of updating hypothesis with new observations. Qwen3-235B~\cite{yang2025qwen3} achieves lower overall performance compared to Gemini-3.1-pro~\cite{Gemini31Pro2026} and GPT-5.4~\cite{OpenAIGPT54_2026}, suggesting that weaker instruction-following reduces the benefit of structured memory on tasks requiring precise categorical judgment.

\begin{figure}[ht]
\centering
\includegraphics[width=\linewidth]{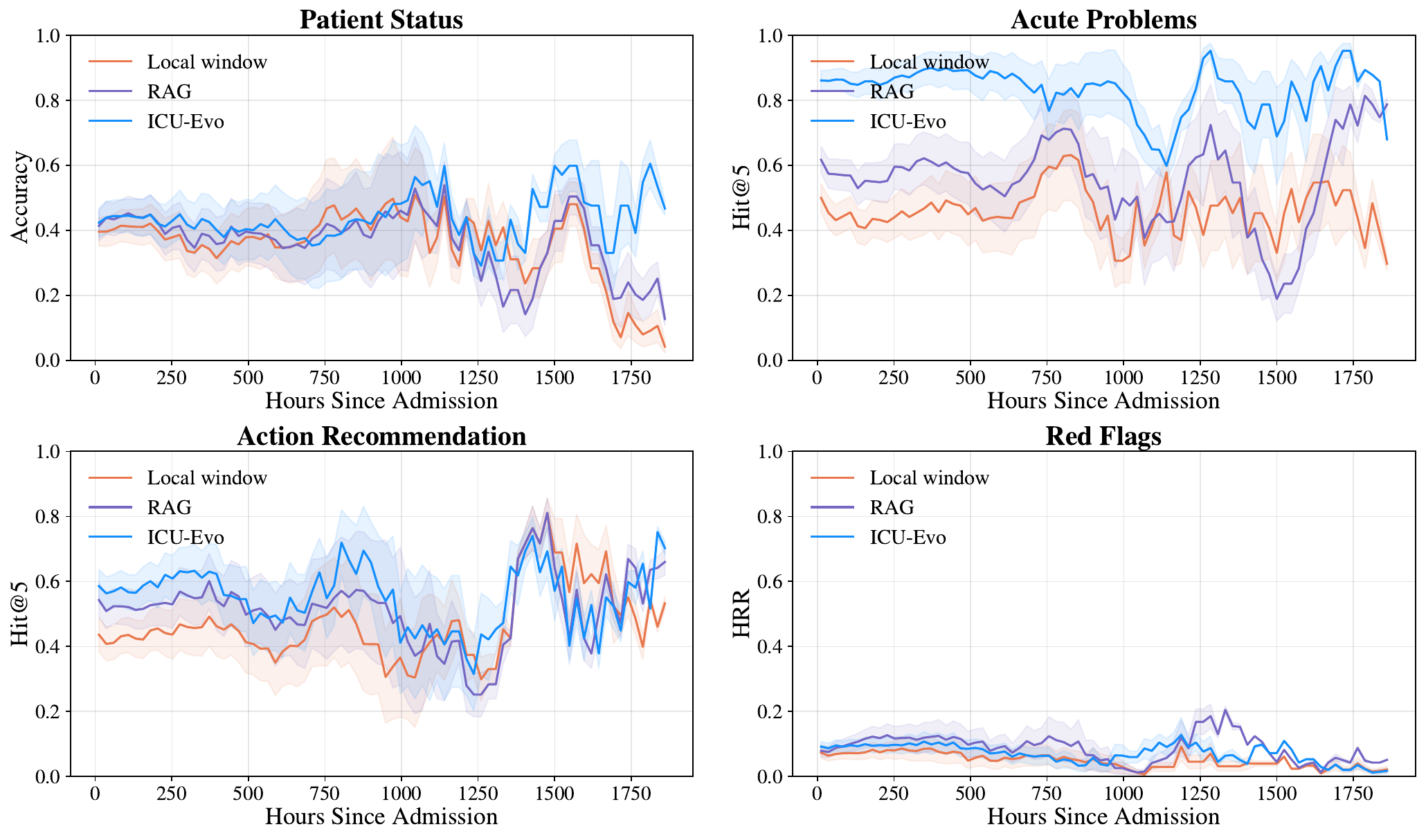}
\caption{Temporal performance over the full ICU stay on \emph{RealICU-Scale} (GPT-5.4~\cite{OpenAIGPT54_2026}).  }
\label{fig:realicu-scale-gpt}
\end{figure}

\begin{figure}[ht]
\centering
\includegraphics[width=\linewidth]{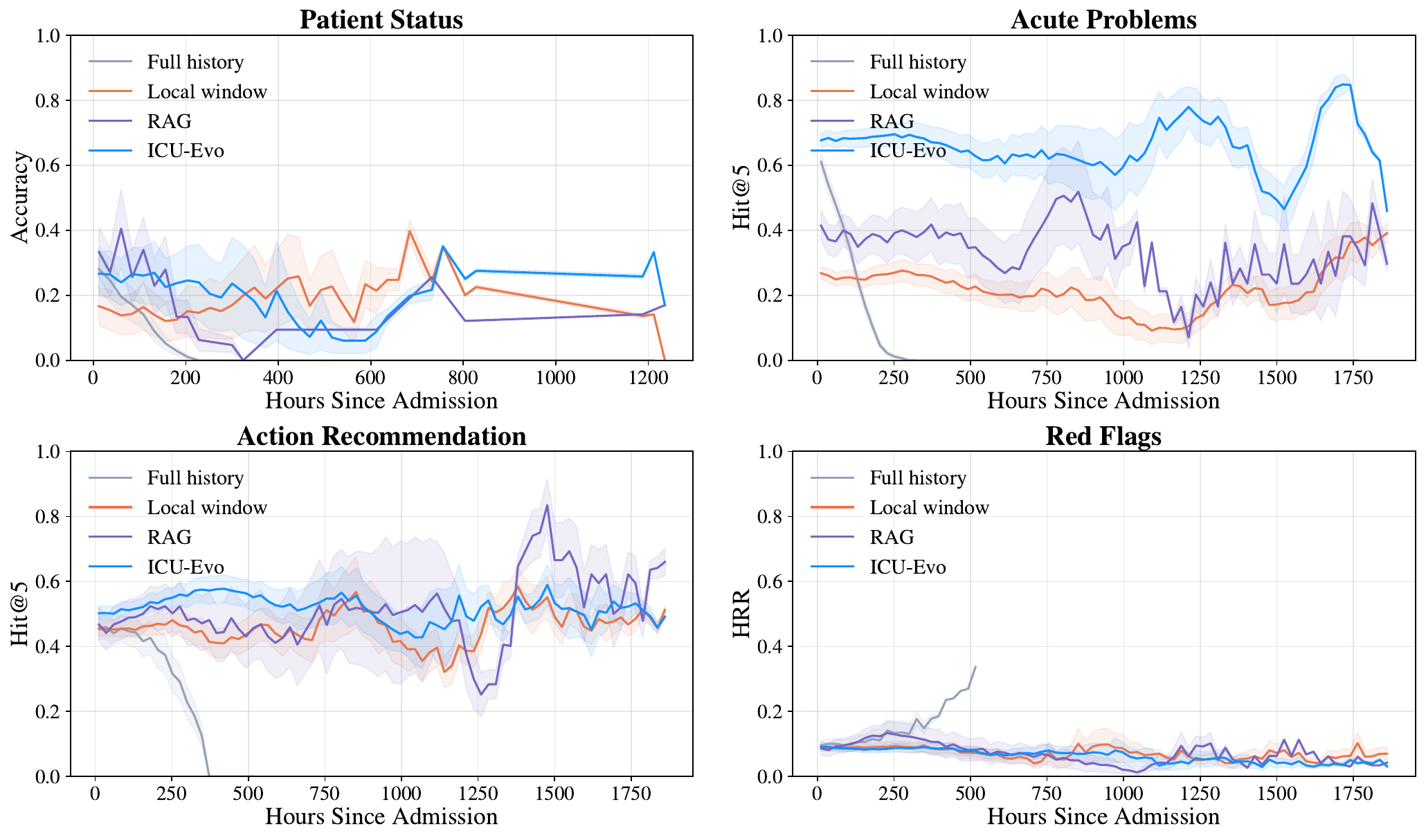}
\caption{Temporal performance over the full ICU stay on \emph{RealICU-Scale} (Qwen3-235B~\cite{yang2025qwen3}). }
\label{fig:realicu-scale-qwen}
\end{figure}

We further illustrate the temporal performance of LLM agents with full-context, local-window, retrieval-augmentation, memory-augmentation configurations in Figure~\ref{fig:realicu-scale-gpt} and Figure~\ref{fig:realicu-scale-qwen}.

\begin{table*}[h]
\centering
\setlength{\tabcolsep}{5pt}
\renewcommand{\arraystretch}{1.18}
\caption{Evaluation results on the \emph{RealICU-Scale}.
Within each backbone, \textbf{bold} marks the best system per column and \underline{underline} the second best.
}
\label{tab:main-results-full}
\small
\begin{tabular}{ll cc cc cc c}
\toprule
 & & \multicolumn{2}{c}{\textbf{\textit{Patient Status}}}
   & \multicolumn{2}{c}{\textbf{\textit{Acute Problems}}}
   & \multicolumn{2}{c}{\textbf{\textit{Action Recom.}}}
   & \textbf{\textit{Red Flags}} \\
\cmidrule(lr){3-4} \cmidrule(lr){5-6} \cmidrule(lr){7-8} \cmidrule(lr){9-9}
\textbf{Backbone} & \textbf{System}
  & Acc.$\uparrow$ & F1$\uparrow$
  & Hit@5$\uparrow$ & R@5$\uparrow$
  & Hit@5$\uparrow$ & R@5$\uparrow$
  & HRR$\downarrow$ \\
\midrule
\multirow{4}{*}{\textbf{Gemini-3.1-pro~\cite{Gemini31Pro2026}}}
  & Full-context  & --     & --     & --     & --     & --     & --     & --     \\
  & Local-window  & 0.405  & \underline{0.264}  & 0.487  & 0.265  & \underline{0.447}  & \underline{0.307}  & \textbf{0.066} \\
  & RAG           & \underline{0.442}  & \underline{0.312}  & \underline{0.568}  & \underline{0.315}  & \underline{0.466}  & \underline{0.331}  & \underline{0.073} \\
\rowcolor{rowhl}
  & ICU-Evo       & \textbf{0.519}  & \textbf{0.348}  & \textbf{0.827}  & \textbf{0.518}  & \textbf{0.514}  & \textbf{0.330}\footnotemark[1]  & 0.087  \\
\midrule
\multirow{4}{*}{\textbf{GPT-5.4~\cite{OpenAIGPT54_2026}}}
  & Full-context  & --     & --     & --     & --     & --     & --     & --     \\
  & Local-window  & \underline{0.415}  & \underline{0.265}  & 0.475  & 0.266  & 0.451  & 0.308  & \textbf{0.073} \\
  & RAG           & 0.411  & 0.269  & \underline{0.584}  & \underline{0.321}  & \underline{0.509}  & \textbf{0.435}  & 0.096  \\
\rowcolor{rowhl}
  & ICU-Evo       & \textbf{0.438}  & \textbf{0.327}  & \textbf{0.852}  & \textbf{0.562}  & \textbf{0.575}  & \underline{0.368}  & \underline{0.090}  \\
\midrule
\multirow{4}{*}{\textbf{Qwen3-235B~\cite{yang2025qwen3}}}
  & Full-context  & 0.201  & 0.116  & \underline{0.401}  & \underline{0.232}  & \underline{0.455}  & 0.299  & \underline{0.215} \\
  & Local-window  & 0.175  & 0.159  & 0.254  & 0.142  & 0.440  & 0.295  & \textbf{0.207} \\
  & RAG           & \textbf{0.367}  & \textbf{0.282}  & 0.379  & 0.207  & 0.446  & \textbf{0.342}  & 0.225  \\
\rowcolor{rowhl}
  & ICU-Evo       & \underline{0.304}  & \underline{0.177}  & \textbf{0.649}  & \textbf{0.375}  & \textbf{0.515}  & \underline{0.327}  & 0.292  \\
\bottomrule
\end{tabular}
\end{table*}

\newpage

\clearpage

\subsection{Averaged Patient Trajectory on \emph{RealICU-Scale}}
\label{sec:app_averaged_trajectory}

\begin{figure}[ht]
\centering
\includegraphics[width=\linewidth]{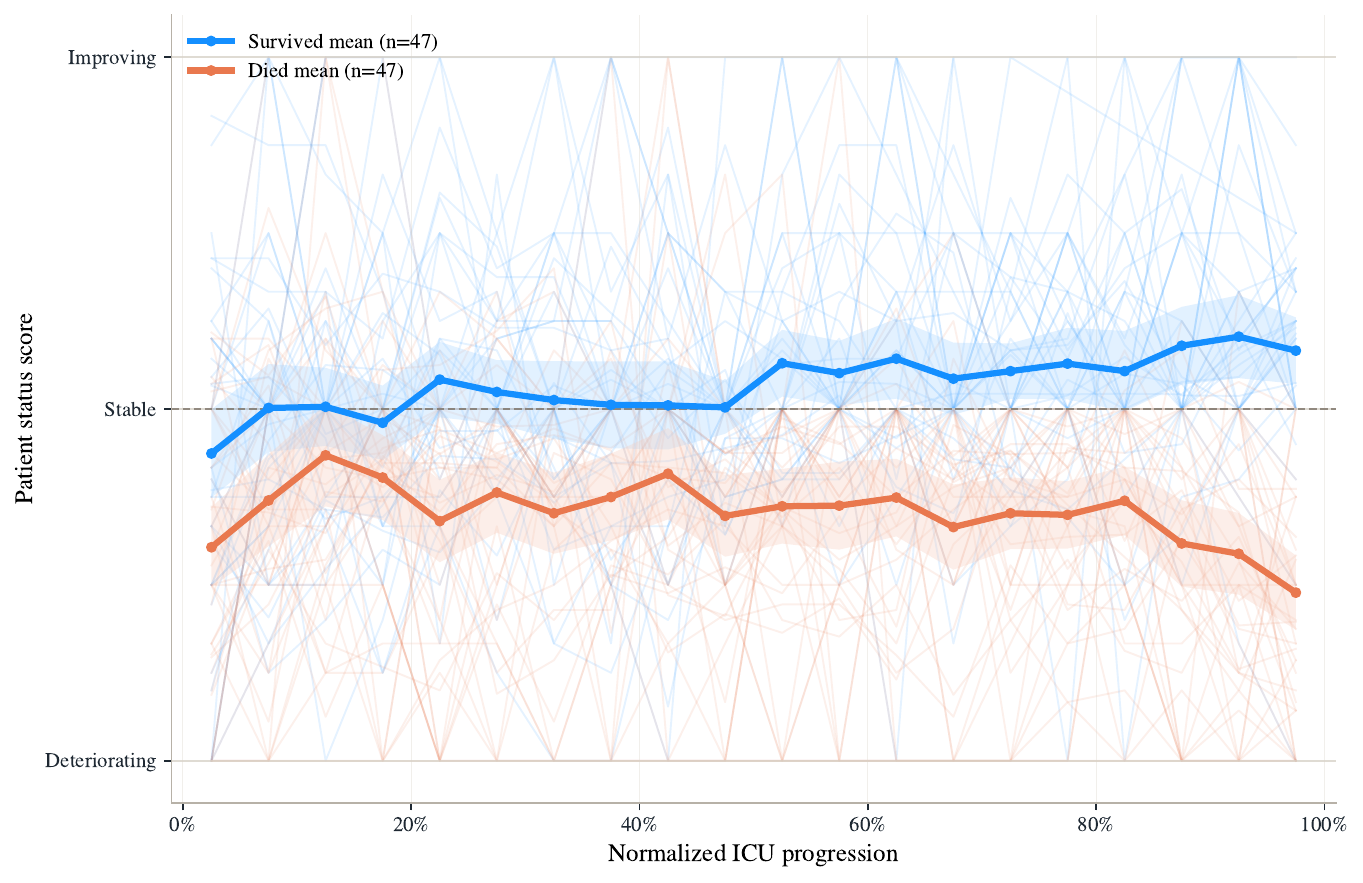}
\caption{Averaged patient status trajectories from \emph{Oracle} on \emph{RealICU-Scale}. Window-level \emph{Patient Status} labels are mapped to an ordinal score (deteriorating $=-1$, stable $=0$, improving $=+1$) and aggregated into normalized duration time. Thin lines show individual trajectories, and thick lines and shaded regions show cohort means with 95\% confidence bands. The survived and died cohorts separate from admission onward and diverge further over the stay.}
\label{fig:oracle-averaged-trajectory}
\end{figure}

We visualize patient trajectories on \emph{RealICU-Scale} using the \emph{Patient Status} label in Figure~\ref{fig:oracle-averaged-trajectory}. We map each window-level label to an ordinal score (deteriorating = -1, stable = 0, improving = 1) and normalize time within each ICU stay to the interval [0,1]. After binning each trajectory into 20 normalized time bins and averaging repeated observations within patient-bin pairs, we plotted all individual trajectories as low-opacity curves and overlaid outcome-stratified cohort means with 95\% confidence bands. This highlights both patient-level heterogeneity and the average temporal separation between survivors and non-survivors.

The survived and died cohorts are already separated at admission, with survivors hovering near \texttt{stable} and non-survivors sitting consistently below it, and the gap widens over the course of the stay as the survivor mean drifts toward \texttt{improving} while the non-survivor mean declines sharply in the final 20\% of normalized ICU time. Both cohorts show substantial patient-level heterogeneity in the thin lines, which is expected given the diversity of admission diagnoses, but the cohort means recover the clinically intuitive ordering that survivors trend upward and non-survivors trend downward. This pattern indicates that the window-level labels produced by \emph{Oracle} aggregate into a coherent patient-level signal, and supports the use of \emph{RealICU-Scale} for trajectory-level analyses despite its labels being generated rather than physician-annotated.

\clearpage

\subsection{Per-Disease Performance on \emph{RealICU-GOLD}.}
\label{sec:app_per_disease}

Tables~\ref{tab:per-disease-gemini} report a breakdown across the six disease groups exceeding 8\% prevalence in \emph{RealICU-Gold} using Gemini-3.1-pro~\cite{Gemini31Pro2026} backbone. The decomposition tests whether the gains of ICU-Evo are driven by a subset of phenotypes or hold across the case mix.

The dominance of ICU-Evo on context-heavy tasks (Patient Status, Acute Problems, Recommended Actions) is consistent across nearly all disease groups. ICU-Evo achieves the best Hit@5 on Acute Problems for every group, with margins over the strongest baseline ranging from 0.134 (GI \& Hepatic) to 0.358 (Sepsis \& Infection), so the benefit of structured longitudinal memory is not phenotype-specific. The pattern is most pronounced on Cardiovascular and GI \& Hepatic cases, where ICU-Evo wins on six of seven metrics, suggesting that diseases with protracted trajectories benefit most from explicit trend and trajectory memory.

Respiratory and Sepsis \& Infection expose the limits of the current memory design on Recommended Actions. RAG matches or surpasses ICU-Evo on Hit@5 and R@5 for these two groups, while ICU-Evo retains its lead on upstream tasks. Respiratory and septic management is dominated by recurring, protocol-driven interventions such as ventilator adjustments and antimicrobial escalation, for which lexical retrieval over recent context is competitive with longitudinal memory. This aligns with the backbone-dependent memory tolerance reported in the Qwen ablation and reinforces that memory architecture and task structure interact.

We report more detailed per-disease results with GPT-5.4~\cite{OpenAIGPT54_2026} backbone in Table \ref{tab:per-disease-gpt}, and with Qwen-235B~\cite{yang2025qwen3} backbone in Table \ref{tab:per-disease-qwen}.

\begin{table*}[h]
\centering
\setlength{\tabcolsep}{5pt}
\renewcommand{\arraystretch}{1.18}
\caption{Per-disease performance on \emph{RealICU-GOLD} (Gemini-3.1-pro~\cite{Gemini31Pro2026} backbone). Disease groups are omitted with less than 8\% proportion. Within each group, \textbf{bold} marks the best system per column and \underline{underline} the second best.}
\label{tab:per-disease-gemini}
\small
\begin{tabular}{ll cc cc cc c}
\toprule
 & & \multicolumn{2}{c}{\textbf{\textit{Patient Status}}}
   & \multicolumn{2}{c}{\textbf{\textit{Acute Problems}}}
   & \multicolumn{2}{c}{\textbf{\textit{Action Recom.}}}
   & \textbf{\textit{Red Flags}} \\
\cmidrule(lr){3-4}\cmidrule(lr){5-6}\cmidrule(lr){7-8}\cmidrule(lr){9-9}
\textbf{Disease Group} & \textbf{System}
  & Acc.$\uparrow$ & F1$\uparrow$
  & Hit@5$\uparrow$ & R@5$\uparrow$
  & Hit@5$\uparrow$ & R@5$\uparrow$
  & HRR@5$\downarrow$ \\
\midrule
\multirow{4}{*}{\textbf{Cardiovascular}}
  & Full-context  & 0.337          & 0.269          & 0.584          & \underline{0.367} & 0.299          & 0.177          & \underline{0.066} \\
  & Local-window  & 0.313          & 0.246          & 0.450          & 0.245          & 0.400          & 0.273          & \textbf{0.059} \\
  & RAG           & \underline{0.409} & \underline{0.370} & \underline{0.593} & 0.337          & \underline{0.442} & \underline{0.284} & \underline{0.066} \\
\rowcolor{rowhl}
  & ICU-Evo       & \textbf{0.472} & \textbf{0.374} & \textbf{0.828} & \textbf{0.526} & \textbf{0.595} & \textbf{0.375} & 0.089 \\
\midrule
\multirow{4}{*}{\textbf{Sepsis \& Infection}}
  & Full-context  & 0.147          & 0.166          & 0.287          & 0.154          & 0.156          & 0.075          & \textbf{0.065} \\
  & Local-window  & 0.327          & 0.268          & 0.458          & 0.258          & 0.378          & 0.226          & \underline{0.067} \\
  & RAG           & \underline{0.353} & \underline{0.283} & \underline{0.574} & \underline{0.312} & \textbf{0.535} & \textbf{0.329} & 0.089 \\
\rowcolor{rowhl}
  & ICU-Evo       & \textbf{0.453} & \textbf{0.335} & \textbf{0.816} & \textbf{0.491} & \underline{0.514} & \underline{0.296} & 0.098 \\
\midrule
\multirow{4}{*}{\textbf{Injury \& Poisoning}}
  & Full-context  & 0.303          & 0.199          & 0.469          & 0.338          & 0.225          & 0.145          & \textbf{0.057} \\
  & Local-window  & 0.303          & 0.209          & \underline{0.512} & 0.306          & 0.390          & 0.257          & 0.090 \\
  & RAG           & \underline{0.474} & \underline{0.379} & 0.636          & \underline{0.369} & \textbf{0.526} & \textbf{0.361} & \underline{0.078} \\
\rowcolor{rowhl}
  & ICU-Evo       & \textbf{0.490} & \textbf{0.399} & \textbf{0.817} & \textbf{0.534} & \underline{0.451} & \underline{0.299} & 0.095 \\
\midrule
\multirow{4}{*}{\textbf{Respiratory}}
  & Full-context  & \textbf{0.340} & 0.248          & \underline{0.621} & \underline{0.390} & 0.379          & 0.207          & 0.112 \\
  & Local-window  & 0.270          & 0.218          & 0.456          & 0.255          & 0.425          & 0.241          & \textbf{0.056} \\
  & RAG           & 0.290          & \underline{0.260} & 0.594          & 0.340          & \underline{0.505} & \underline{0.284} & \underline{0.084} \\
\rowcolor{rowhl}
  & ICU-Evo       & \underline{0.320} & \textbf{0.278} & \textbf{0.851} & \textbf{0.559} & \textbf{0.551} & \textbf{0.323} & 0.120 \\
\midrule
\multirow{4}{*}{\textbf{GI \& Hepatic}}
  & Full-context  & \underline{0.438} & \underline{0.396} & 0.527          & 0.338          & 0.347          & 0.226          & 0.115 \\
  & Local-window  & 0.350          & 0.311          & 0.510          & 0.298          & 0.427          & 0.317          & 0.122 \\
  & RAG           & 0.412          & 0.391          & \underline{0.645} & \underline{0.408} & \underline{0.587} & \underline{0.360} & \textbf{0.073} \\
\rowcolor{rowhl}
  & ICU-Evo       & \textbf{0.500} & \textbf{0.437} & \textbf{0.779} & \textbf{0.512} & \textbf{0.592} & \textbf{0.382} & \underline{0.098} \\
\midrule
\multirow{4}{*}{\textbf{All Diseases (Table~\ref{tab:main-results-gold})}}
  & Full-context           & 0.298 & 0.258          & 0.486          & 0.308          & 0.259          & 0.152          & \textbf{0.137} \\
  & Local-window           & 0.315 & 0.239          & 0.459          & 0.258          & 0.395          & 0.260          & \underline{0.151} \\
  & RAG                    & \underline{0.402} & \underline{0.348} & \underline{0.596} & \underline{0.342} & \underline{0.496} & \underline{0.313} & 0.216 \\
\rowcolor{rowhl}
  & ICU-Evo       & \textbf{0.459} & \textbf{0.365} & \textbf{0.823} & \textbf{0.526} & \textbf{0.676} & \textbf{0.534} & 0.300 \\
\bottomrule
\end{tabular}
\end{table*}

\clearpage

\begin{table*}[h]
\centering
\setlength{\tabcolsep}{5pt}
\renewcommand{\arraystretch}{1.18}
\caption{Per-disease performance on \emph{RealICU-GOLD} (GPT-5.4~\cite{OpenAIGPT54_2026} backbone). Disease groups are omitted with less than 8\% proportion. Within each group, \textbf{bold} marks the best system per column and \underline{underline} the second best.}
\label{tab:per-disease-gpt}
\small
\begin{tabular}{ll cc cc cc c}
\toprule
 & & \multicolumn{2}{c}{\textbf{\textit{Patient Status}}}
   & \multicolumn{2}{c}{\textbf{\textit{Acute Problems}}}
   & \multicolumn{2}{c}{\textbf{\textit{Action Recom.}}}
   & \textbf{\textit{Red Flags}} \\
\cmidrule(lr){3-4}\cmidrule(lr){5-6}\cmidrule(lr){7-8}\cmidrule(lr){9-9}
\textbf{Disease Group} & \textbf{System}
  & Acc.$\uparrow$ & F1$\uparrow$
  & Hit@5$\uparrow$ & R@5$\uparrow$
  & Hit@5$\uparrow$ & R@5$\uparrow$
  & HRR@5$\downarrow$ \\
\midrule
\multirow{4}{*}{\textbf{Cardiovascular}}
  & Full-context  & \textbf{0.350} & \textbf{0.250} & \underline{0.589} & \underline{0.396} & \underline{0.518} & \underline{0.387} & 0.157 \\
  & Local-window  & 0.262          & 0.190          & 0.484          & 0.286          & 0.327          & 0.297          & \textbf{0.114} \\
  & RAG           & \underline{0.300} & 0.243       & 0.578          & 0.313          & 0.487          & 0.456          & 0.138 \\
\rowcolor{rowhl}
  & ICU-Evo       & 0.275          & \underline{0.246} & \textbf{0.853} & \textbf{0.558} & \textbf{0.705} & \textbf{0.562} & \underline{0.136} \\
\midrule
\multirow{4}{*}{\textbf{Sepsis \& Infection}}
  & Full-context  & 0.153          & 0.164          & 0.314          & 0.191          & 0.137          & 0.087          & 0.111 \\
  & Local-window  & 0.153          & 0.213          & 0.528          & 0.293          & 0.368          & 0.233          & \underline{0.114} \\
  & RAG           & \underline{0.227} & \textbf{0.292} & \underline{0.609} & \underline{0.329} & \underline{0.430} & \underline{0.327} & \textbf{0.095} \\
\rowcolor{rowhl}
  & ICU-Evo       & \textbf{0.347} & \underline{0.266} & \textbf{0.864} & \textbf{0.539} & \textbf{0.700} & \textbf{0.558} & 0.158 \\
\midrule
\multirow{4}{*}{\textbf{Injury \& Poisoning}}
  & Full-context  & 0.249          & 0.157          & 0.510          & \underline{0.373} & 0.404          & 0.306          & 0.161 \\
  & Local-window  & \underline{0.319} & 0.150       & 0.479          & 0.299          & 0.482          & 0.318          & 0.138 \\
  & RAG           & \textbf{0.328} & \underline{0.182} & \underline{0.545} & 0.364          & \underline{0.545} & \underline{0.431} & 0.157 \\
\rowcolor{rowhl}
  & ICU-Evo       & \textbf{0.328} & \textbf{0.267} & \textbf{0.922} & \textbf{0.648} & \textbf{0.618} & \textbf{0.455} & \textbf{0.104} \\
\midrule
\multirow{4}{*}{\textbf{Respiratory}}
  & Full-context  & \underline{0.340} & \underline{0.254} & \underline{0.653} & \underline{0.438} & \underline{0.543} & \underline{0.477} & 0.176 \\
  & Local-window  & 0.322          & 0.189          & 0.533          & 0.297          & 0.386          & 0.211          & \underline{0.137} \\
  & RAG           & \textbf{0.378} & \textbf{0.290} & 0.678          & 0.424          & 0.319          & 0.247          & \textbf{0.113} \\
\rowcolor{rowhl}
  & ICU-Evo       & 0.270          & 0.237          & \textbf{0.923} & \textbf{0.631} & \textbf{0.767} & \textbf{0.623} & 0.182 \\
\midrule
\multirow{4}{*}{\textbf{GI \& Hepatic}}
  & Full-context  & \textbf{0.425} & \textbf{0.349} & 0.575          & 0.441          & 0.474          & 0.323          & 0.306 \\
  & Local-window  & 0.100          & 0.165          & 0.561          & 0.373          & 0.331          & 0.279          & \textbf{0.098} \\
  & RAG           & 0.214          & 0.256          & \underline{0.682} & \underline{0.512} & \underline{0.512} & \underline{0.412} & \underline{0.148} \\
\rowcolor{rowhl}
  & ICU-Evo       & \underline{0.363} & \underline{0.347} & \textbf{0.827} & \textbf{0.570} & \textbf{0.657} & \textbf{0.490} & 0.157 \\
\midrule
\multirow{4}{*}{\textbf{All Diseases (Table~\ref{tab:main-results-gold})}}
  & Full-context           & \underline{0.294} & 0.233 & 0.510 & 0.348 & 0.404 & 0.300 & 0.298 \\
  & Local-window           & 0.233 & 0.184 & 0.500 & 0.293 & 0.380 & 0.281 & \textbf{0.165} \\
  & RAG                    & 0.288 & \underline{0.256} & \underline{0.599} & \underline{0.349} & \underline{0.480} & \underline{0.398} & \underline{0.234} \\
\rowcolor{rowhl}
  & ICU-Evo       & \textbf{0.312} & \textbf{0.264} & \textbf{0.867} & \textbf{0.570} & \textbf{0.676} & \textbf{0.534} & 0.473 \\
\bottomrule
\end{tabular}
\end{table*}

\clearpage

\begin{table*}[h]
\centering
\setlength{\tabcolsep}{5pt}
\renewcommand{\arraystretch}{1.18}
\caption{Per-disease performance on \emph{RealICU-GOLD} (Qwen3-235B~\cite{yang2025qwen3} backbone). Disease groups are omitted with less than 8\% proportion. Within each group, \textbf{bold} marks the best system per column and \underline{underline} the second best.}
\label{tab:per-disease-qwen}
\small
\begin{tabular}{ll cc cc cc c}
\toprule
 & & \multicolumn{2}{c}{\textbf{\textit{Patient Status}}}
   & \multicolumn{2}{c}{\textbf{\textit{Acute Problems}}}
   & \multicolumn{2}{c}{\textbf{\textit{Action Recom.}}}
   & \textbf{\textit{Red Flags}} \\
\cmidrule(lr){3-4}\cmidrule(lr){5-6}\cmidrule(lr){7-8}\cmidrule(lr){9-9}
\textbf{Disease Group} & \textbf{System}
  & Acc.$\uparrow$ & F1$\uparrow$
  & Hit@5$\uparrow$ & R@5$\uparrow$
  & Hit@5$\uparrow$ & R@5$\uparrow$
  & HRR@5$\downarrow$ \\
\midrule
\multirow{4}{*}{\textbf{Cardiovascular}}
  & Full-context  & 0.218          & 0.206          & \underline{0.455} & 0.249          & 0.390          & 0.270          & 0.129 \\
  & Local-window  & 0.156          & 0.164          & 0.188          & 0.109          & 0.351          & 0.246          & \textbf{0.087} \\
  & RAG           & \textbf{0.307} & \textbf{0.278} & 0.350          & 0.189          & \underline{0.451} & \underline{0.332} & \underline{0.090} \\
\rowcolor{rowhl}
  & ICU-Evo       & \underline{0.268} & \underline{0.219} & \textbf{0.552} & \textbf{0.316} & \textbf{0.552} & \textbf{0.363} & 0.134 \\
\midrule
\multirow{4}{*}{\textbf{Sepsis \& Infection}}
  & Full-context  & 0.107          & 0.122          & 0.233          & 0.137          & 0.156          & 0.108          & \textbf{0.083} \\
  & Local-window  & \underline{0.147} & \underline{0.186} & 0.224          & 0.113          & 0.322          & 0.198          & 0.098 \\
  & RAG           & \textbf{0.333} & \textbf{0.282} & \underline{0.348} & \underline{0.176} & \underline{0.443} & \underline{0.277} & \underline{0.093} \\
\rowcolor{rowhl}
  & ICU-Evo       & 0.140          & 0.130          & \textbf{0.575} & \textbf{0.328} & \textbf{0.530} & \textbf{0.353} & 0.098 \\
\midrule
\multirow{4}{*}{\textbf{Injury \& Poisoning}}
  & Full-context  & 0.269          & 0.153          & 0.432          & 0.289          & 0.291          & 0.232          & 0.093 \\
  & Local-window  & 0.197          & 0.165          & 0.301          & 0.221          & 0.373          & 0.280          & \textbf{0.065} \\
  & RAG           & \textbf{0.362} & \textbf{0.252} & \underline{0.516} & \underline{0.328} & \textbf{0.481} & \textbf{0.357} & 0.102 \\
\rowcolor{rowhl}
  & ICU-Evo       & \underline{0.305} & \underline{0.220} & \textbf{0.669} & \textbf{0.457} & \underline{0.452} & \underline{0.321} & \underline{0.087} \\
\midrule
\multirow{4}{*}{\textbf{Respiratory}}
  & Full-context  & 0.280          & 0.145          & 0.444          & 0.266          & 0.419          & 0.236          & 0.116 \\
  & Local-window  & 0.140          & 0.117          & 0.253          & 0.147          & 0.354          & 0.222          & \textbf{0.052} \\
  & RAG           & \underline{0.340} & \textbf{0.311} & \underline{0.451} & \underline{0.272} & \underline{0.463} & \underline{0.331} & \underline{0.083} \\
\rowcolor{rowhl}
  & ICU-Evo       & \textbf{0.380} & \underline{0.205} & \textbf{0.658} & \textbf{0.404} & \textbf{0.675} & \textbf{0.457} & 0.162 \\
\midrule
\multirow{4}{*}{\textbf{GI \& Hepatic}}
  & Full-context  & \textbf{0.375} & \underline{0.315} & \underline{0.459} & \underline{0.295} & 0.382          & 0.270          & 0.188 \\
  & Local-window  & 0.175          & 0.150          & 0.214          & 0.138          & 0.334          & 0.273          & \underline{0.123} \\
  & RAG           & \underline{0.338} & \textbf{0.336} & 0.329          & 0.199          & \underline{0.459} & \underline{0.369} & \textbf{0.099} \\
\rowcolor{rowhl}
  & ICU-Evo       & 0.262          & 0.258          & \textbf{0.646} & \textbf{0.413} & \textbf{0.508} & \textbf{0.394} & 0.132 \\
\midrule
\multirow{4}{*}{\textbf{All Diseases (Table~\ref{tab:main-results-gold})}}
& Full-context           & 0.225 & \underline{0.188} & \underline{0.384} & \underline{0.226} & 0.329 & 0.222 & 0.117 \\
  & Local-window           & 0.152 & 0.154 & 0.213 & 0.126 & 0.352 & 0.242 & \textbf{0.080} \\
  & RAG                    & \textbf{0.315} & \textbf{0.271} & 0.379 & 0.211 & \underline{0.453} & \underline{0.324} & \underline{0.095} \\
\rowcolor{rowhl}
  & ICU-Evo       & \underline{0.253} & 0.197 & \textbf{0.600} & \textbf{0.362} & \textbf{0.526} & \textbf{0.357} & 0.117 \\
\bottomrule
\end{tabular}
\end{table*}

\clearpage

\subsection{Semantic Matcher Calibration}
\label{app:semantic_match}

We adopt PubMedBERT~\cite{gu2021domain} (\texttt{NeuML/pubmedbert-base-embeddings}) to generate embeddings for semantic match for  \emph{Acute Problems}, \emph{Action Recommendation}, and \emph{Red Flags} tasks.

\paragraph{Calibration set.}
We sampled 100 action-string pairs from held-out ICU windows and asked a
board-certified intensivist to label each pair as a binary classification for semantic match or non-match. The set is balanced by construction with 50 matched pairs and
50 non-matched pairs. 

\begin{figure}[ht]
\centering
\includegraphics[width=0.55\linewidth]{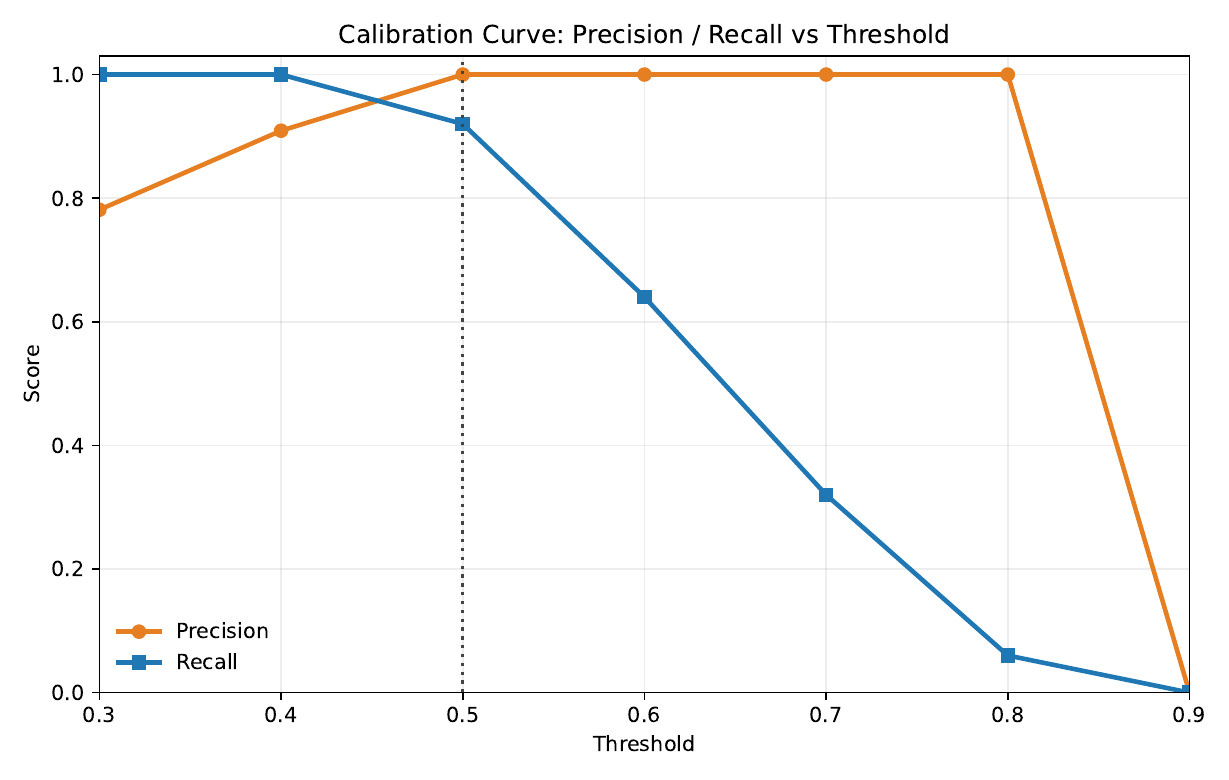}
\caption{Evaluating PubMedBERT~\cite{gu2021domain} matcher on the calibration set under different thresholds. The selected $\tau^{*} = 0.5$ (dashed line) achieves the best overall performance.}
\label{fig:matcher-calibration}
\end{figure}

\paragraph{Threshold sweep.}
Table~\ref{tab:matcher-calibration} reports precision, recall, F1, and accuracy at seven candidate thresholds, and Figure~\ref{fig:matcher-calibration} visualises the trade-off. PubMedBERT cosine similarity separates the two classes almost perfectly (AUROC $= 0.996$). Precision reaches $1.00$ for all $\tau \geq 0.5$, while recall decays monotonically as $\tau$ increases. We select $\tau^{*} = 0.5$, which maximises F1 ($0.958$) and eliminates false positives while retaining $92\%$ of true matches. This operating point is used in all reported evaluations.
 
\begin{table}[ht]
\centering
\small
\caption{Semantic matcher performance on the 100-pair calibration set
across candidate thresholds. The selected threshold
($\tau^{*}=0.5$) maximises F1 and achieves perfect precision.}
\label{tab:matcher-calibration}
\begin{tabular}{lcccc}
\toprule
$\tau$ & Accuracy & Precision & Recall & F1 \\
\midrule
0.3 & 0.86 & 0.78 & 1.00 & 0.88 \\
0.4 & 0.95 & 0.91 & 1.00 & 0.95 \\
\textbf{0.5}$^{*}$ & \textbf{0.96} & \textbf{1.00} & \textbf{0.92} & \textbf{0.96} \\
0.6 & 0.82 & 1.00 & 0.64 & 0.78 \\
0.7 & 0.66 & 1.00 & 0.32 & 0.48 \\
0.8 & 0.53 & 1.00 & 0.06 & 0.12 \\
0.9 & 0.50 & 0.00 & 0.00 & 0.00 \\
\bottomrule
\end{tabular}
\end{table}
 
\clearpage
\subsection{Effect of Trajectory Summarization}
\label{app:trajectory-summary}

To investigate whether ICU-Evo's improvements arise primarily
from trajectory compression, we introduce a Trajectory Summary
+ Local Window baseline using the same summarization module
as ICU-Evo. This baseline combines the summarized patient
history with the current local window, while keeping other
experimental settings unchanged.

As shown in Table~\ref{tab:summary-baseline},
trajectory summarization substantially improves performance
over the local-window baseline, particularly on Acute Problems
and Action Recommendation. ICU-Evo further improves
Acute Problems Hit@5 from 0.732 to 0.823 and Action
Recommendation Hit@5 from 0.525 to 0.676, suggesting
benefits from structured memory beyond trajectory compression.
However, the higher recommendation coverage is accompanied
by an increased HRR of red flag actions, consistent with the recall--safety
trade-off discussed in Section~\ref{sec:eval:failure}

\begin{table}[h]
\centering
\setlength{\tabcolsep}{5pt}
\renewcommand{\arraystretch}{1.15}
\caption{Effect of trajectory summarization on
\emph{RealICU-Gold} using Gemini-3.1-pro~\cite{Gemini31Pro2026}.
\textbf{Bold} marks the best result per column.}
\label{tab:summary-baseline}
\small
\begin{tabular}{lccccccc}
\toprule
& \multicolumn{2}{c}{\textbf{\textit{Patient Status}}}
& \multicolumn{2}{c}{\textbf{\textit{Acute Problems}}}
& \multicolumn{2}{c}{\textbf{\textit{Action Recom.}}}
& \textbf{\textit{Red Flags}} \\
\cmidrule(lr){2-3}
\cmidrule(lr){4-5}
\cmidrule(lr){6-7}
\cmidrule(lr){8-8}
\textbf{System} &
Acc.$\uparrow$ & F1$\uparrow$ &
Hit@5$\uparrow$ & R@5$\uparrow$ &
Hit@5$\uparrow$ & R@5$\uparrow$ &
HRR$\downarrow$ \\
\midrule
Local-window
& 0.315 & 0.239
& 0.459 & 0.258
& 0.395 & 0.260
& 0.151 \\

Summary + Local
& 0.378 & 0.304
& 0.732 & 0.452
& 0.525 & 0.327
& \textbf{0.081} \\

\rowcolor{rowhl}
ICU-Evo
& \textbf{0.459} & \textbf{0.365}
& \textbf{0.823} & \textbf{0.526}
& \textbf{0.676} & \textbf{0.534}
& 0.300 \\
\bottomrule
\end{tabular}
\end{table}

\clearpage
\subsection{Token Efficiency}
\label{app:token-efficiency}

We assess token efficiency from two complementary perspectives, namely the per-prediction cost and the longitudinal coverage delivered per input token. A direct comparison of raw token counts suggests that ICU-Evo is more expensive than the local-window and RAG baselines. This view, however, omits a central design objective of ICU-Evo, which is to surface broad trajectory context at every prediction step. We therefore report a coverage-normalized metric alongside the raw cost.

\paragraph{Per-prediction cost.}
We report the average input and total tokens per prediction on the Qwen run, with RAG projected to match the call volume of the local-window baseline. ICU-Evo is not the cheapest configuration in raw tokens, yet it is substantially cheaper than full-context prompting while remaining more expensive than the local-window and RAG baselines, as shown in Table~\ref{tab:token-cost}.

\paragraph{Coverage-normalized efficiency.}
To account for the trajectory context that each mode actually surfaces, we define the covered windows per prediction as $1$ for the local-window baseline, $1 + k$ for RAG with $k$ retrieved windows, and $\text{window\_index} + 1$ for ICU-Evo and the full-context baseline, reflecting the current window together with all accumulated prior context. We then report the covered windows per million input tokens and the input tokens consumed per covered window on the \emph{Patient Status} task. Once normalized by timeline coverage, ICU-Evo becomes the most input-efficient mode, achieving the highest coverage density and the lowest input-token cost per covered window, as shown in Table~\ref{tab:token-coverage}.

Taken together, these two views indicate that, although ICU-Evo consumes more tokens per prediction than the local-window baseline, it delivers substantially denser longitudinal context per input token, which reflects better token utilization for timeline-aware reasoning in the ICU.

\begin{table}[h]
\centering
\small
\caption{Per-prediction token cost on \emph{RealICU-Scale} with Qwen3-235B~\cite{yang2025qwen3}.}
\label{tab:token-cost}
\begin{tabular}{lrrrr}
\toprule
Mode & Predictions & Input tokens & Avg.\ input / pred. & Avg.\ total / pred. \\
\midrule
Full-context     & 11{,}065 & 405{,}300{,}948 & 36{,}629.10 & 36{,}763.84 \\
Local-window     & 11{,}862 &  25{,}382{,}499 &  2{,}139.82 &  2{,}319.49 \\
RAG (projected)  & 11{,}862 &  83{,}797{,}771 &  7{,}064.39 &  7{,}318.39 \\
\rowcolor{rowhl}
ICU-Evo          & 11{,}862 & 254{,}272{,}587 & 21{,}435.90 & 21{,}971.55 \\
\bottomrule
\end{tabular}
\end{table}

\begin{table}[h]
\centering
\small
\caption{Coverage-normalized input efficiency on Patient Status.}
\label{tab:token-coverage}
\begin{tabular}{lrrr}
\toprule
Mode & Covered windows & Windows / 1M input tok. & Input tok.\ / window \\
\midrule
Full-context  & 5{,}112{,}589 & 12{,}614.30 &    79.28 \\
Local-window  &     11{,}862 &     467.33 & 2{,}139.82 \\
RAG           &      6{,}130 &     541.32 & 1{,}847.34 \\
\rowcolor{rowhl}
ICU-Evo       & 6{,}304{,}410 & 24{,}793.90 &    40.33 \\
\bottomrule
\end{tabular}
\end{table}

\FloatBarrier

%% file: chapters/appendix/dataset_details.tex
\newpage
\section{Dataset Details}
\label{sec:app_dataset_details}

\subsection{Dataset Statistics}
\label{sec:app_dataset_statistics}

\begin{figure}[htbp]
    \centering
    \begin{subfigure}[b]{0.3\textwidth}
        \includegraphics[width=\textwidth]{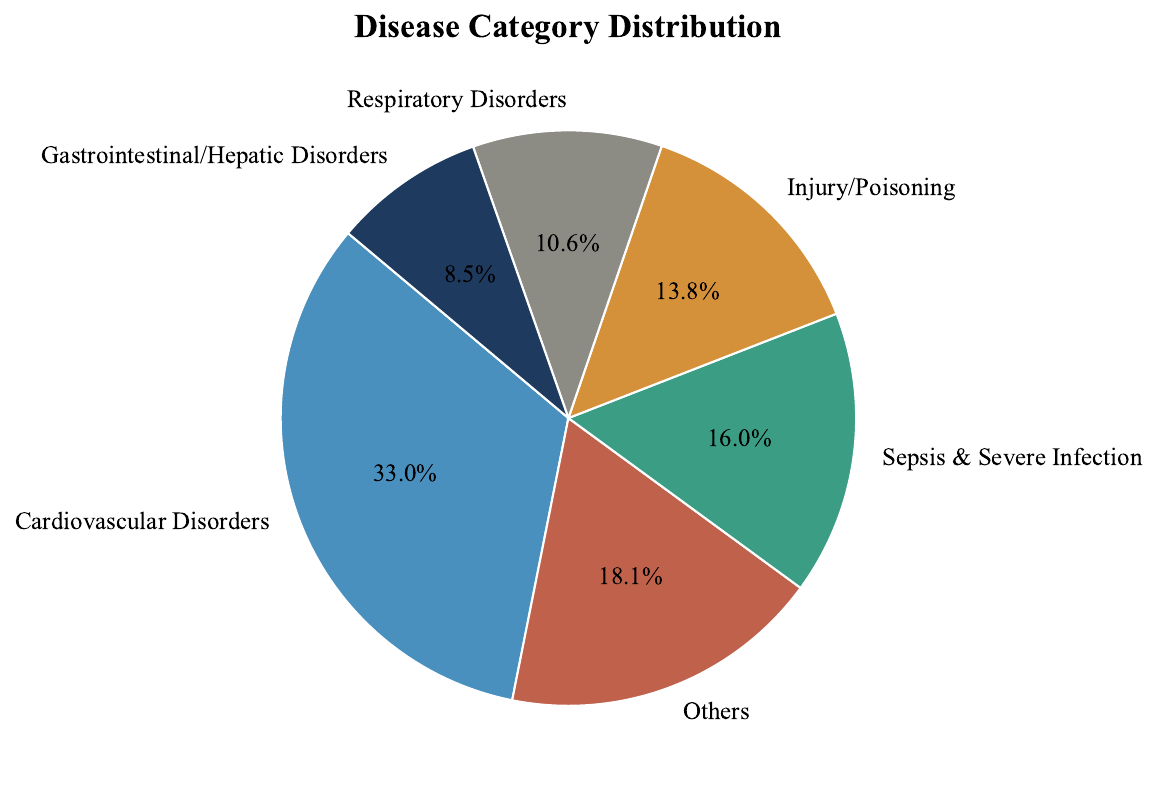}
    \end{subfigure}
    \hfill
    \begin{subfigure}[b]{0.3\textwidth}
        \includegraphics[width=\textwidth]{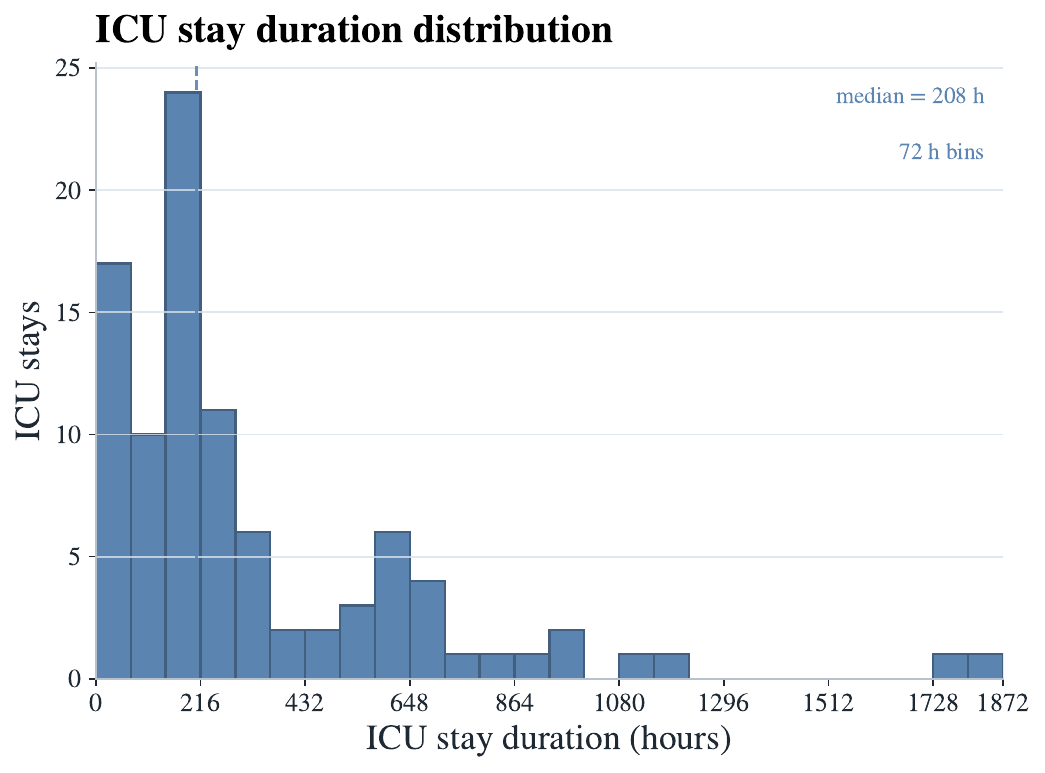}
    \end{subfigure}
    \hfill
    \begin{subfigure}[b]{0.3\textwidth}
        \includegraphics[width=\textwidth]{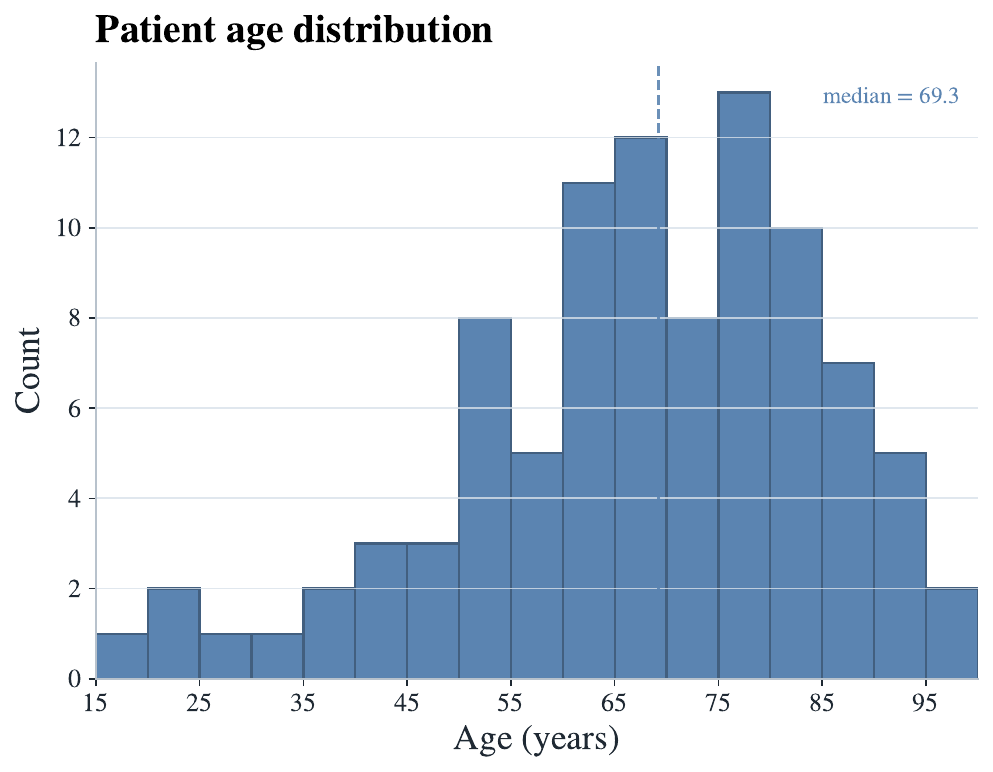}
    \end{subfigure}

    \vspace{0.5em}

    \begin{subfigure}[b]{0.3\textwidth}
        \includegraphics[width=\textwidth]{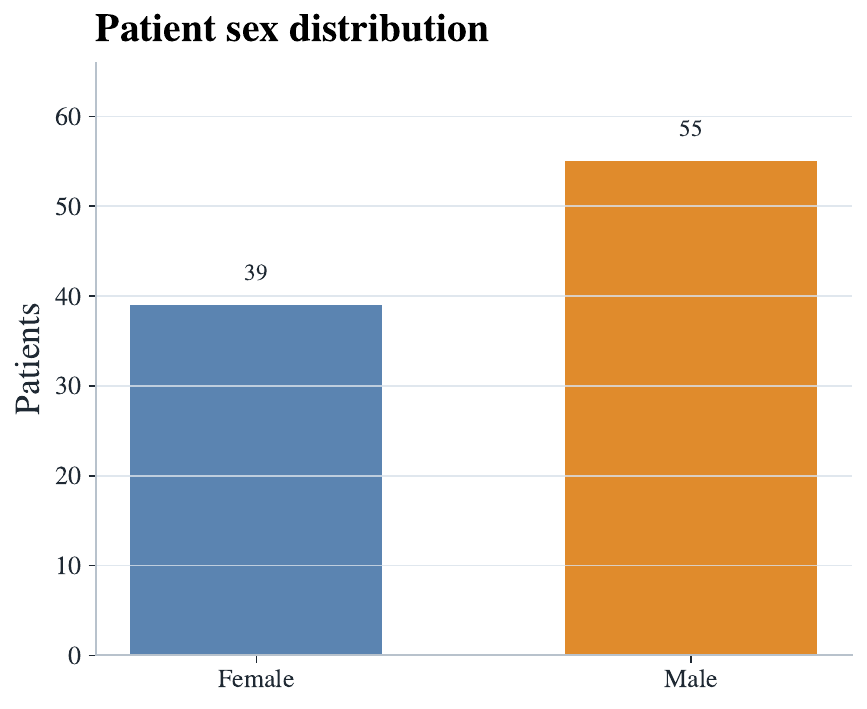}
    \end{subfigure}
    \hfill
    \begin{subfigure}[b]{0.3\textwidth}
        \includegraphics[width=\textwidth]{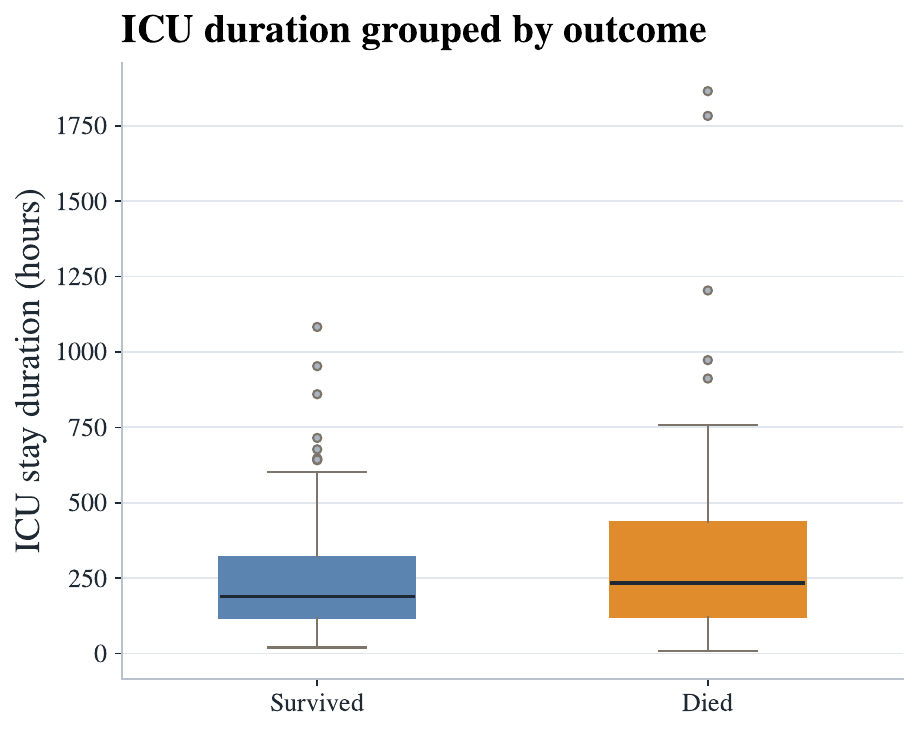}
    \end{subfigure}
    \hfill
    \begin{subfigure}[b]{0.3\textwidth}
        \includegraphics[width=\textwidth]{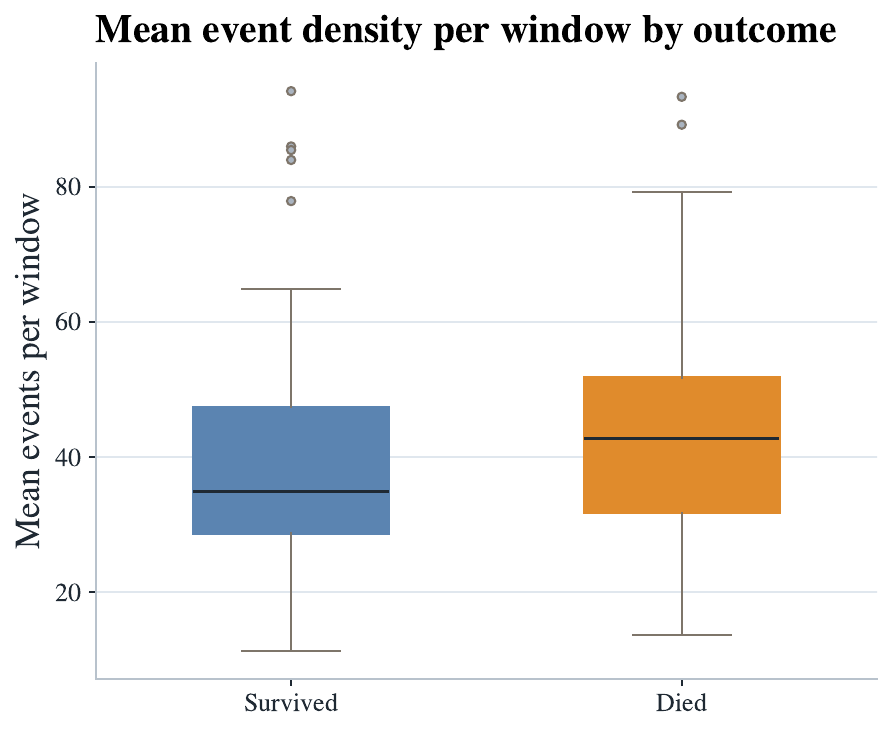}
    \end{subfigure}

    \caption{Cohort Demographics and Clinical Characteristics of the 94-Patient \emph{RealICU} Cohort}
    \label{fig:cohort_stats}
\end{figure}

\paragraph{Cohort Statistics}
Figure~\ref{fig:cohort_stats} summarizes the demographic and clinical composition of the selected 94-patient cohort across six dimensions: disease category, ICU stay duration, age, sex, stay duration stratified by survival outcome, and mean event density per window stratified by outcome. 

To categorize each ICU stay with disease types, we extract the diagnosis code closest in time to ICU admission. ICD-9 and ICD-10 codes were then mapped to broad disease categories using a rule-based grouping based on ICD chapters, with sepsis-related codes grouped into a dedicated Sepsis and Severe Infection category. Specifically, the largest disease group was Cardiovascular Disorders (32.98\%), followed by Sepsis and Severe Infection (15.96\%), Injury/Poisoning (13.83\%), Respiratory Disorders (10.64\%), and Gastrointestinal/Hepatic Disorders (8.51\%). The remaining categories were less common: Neurological Disorders (4.26\%); Clinical Signs/Symptoms, Congenital Disorders, Infectious Diseases, and Oncology (2.13\% each); and Endocrine/Metabolic Disorders, Hematologic Disorders, Musculoskeletal Disorders, Psychiatric Disorders, and Renal/Genitourinary Disorders (1.06\% each). In the pie chart, disease categories below 5\% were merged into Others

\begin{figure}[htbp]
    \centering
    \begin{subfigure}[b]{0.50\textwidth}
        \includegraphics[width=\textwidth]{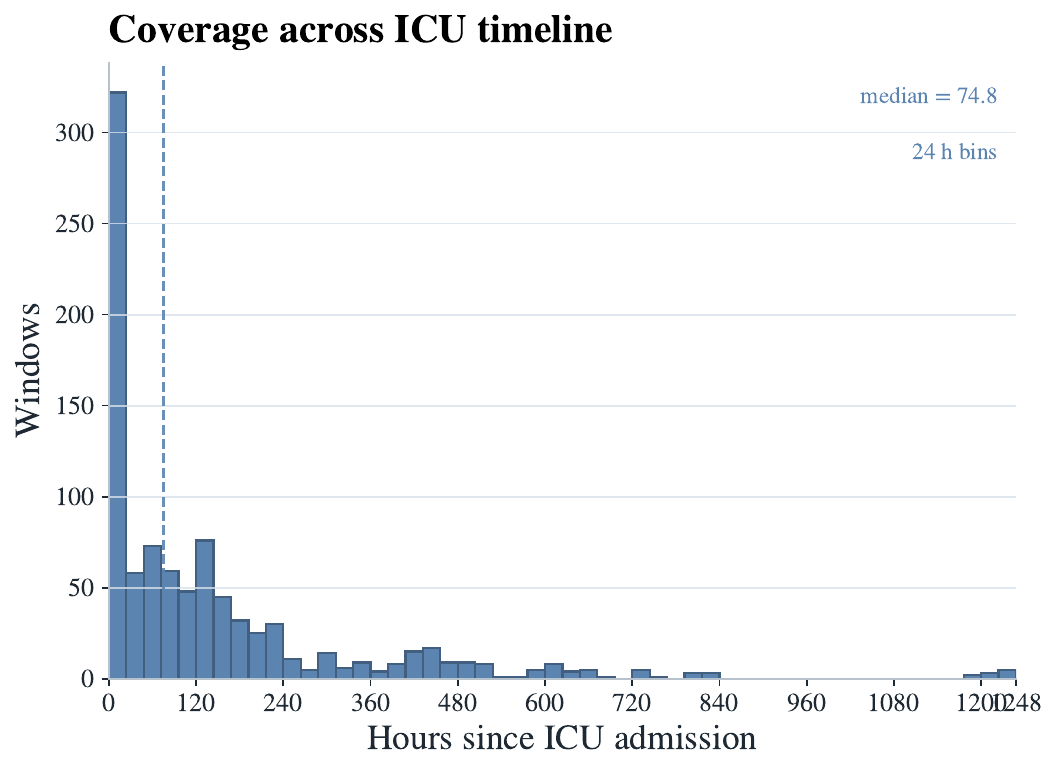}
    \end{subfigure}
    \hfill
    \begin{subfigure}[b]{0.35\textwidth}
        \includegraphics[width=\textwidth]{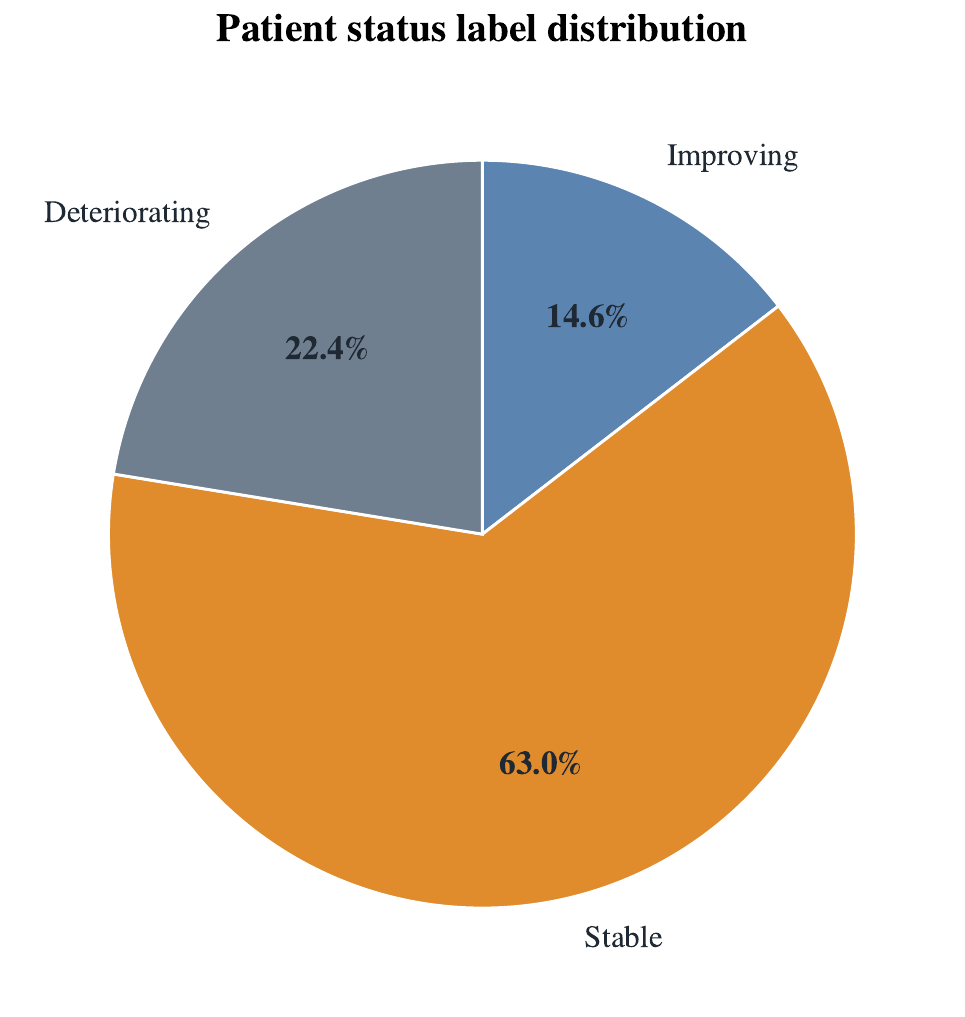}
    \end{subfigure}

    \vspace{0.5em}

    \begin{subfigure}[b]{0.3\textwidth}
        \includegraphics[width=\textwidth]{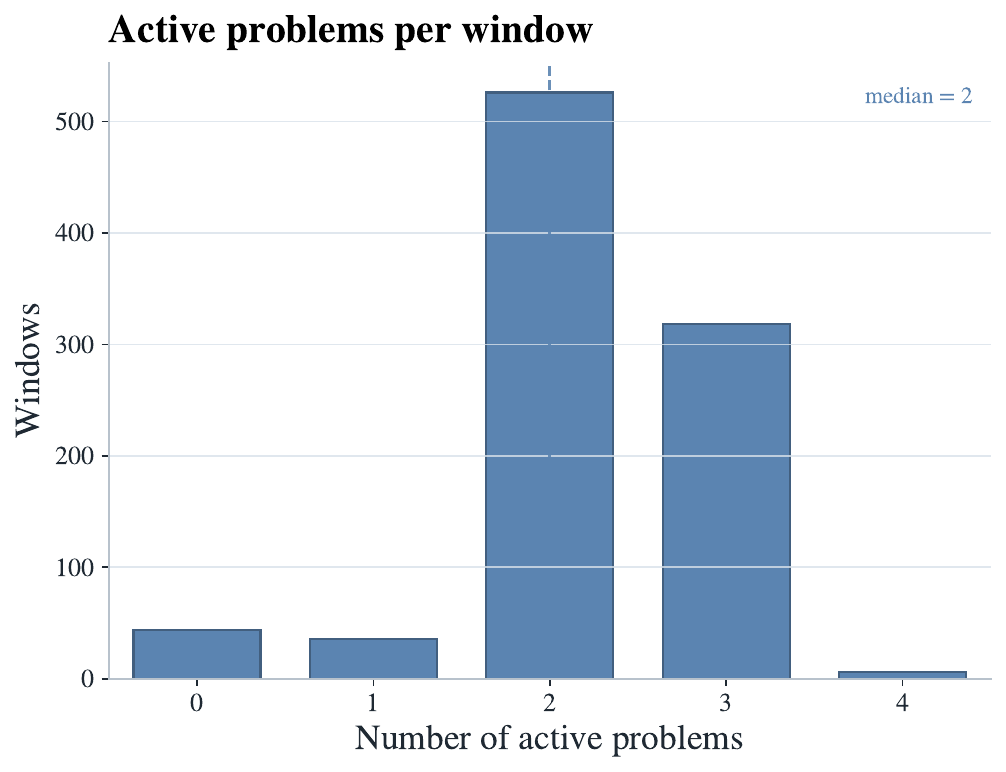}
    \end{subfigure}
    \hfill
    \begin{subfigure}[b]{0.3\textwidth}
        \includegraphics[width=\textwidth]{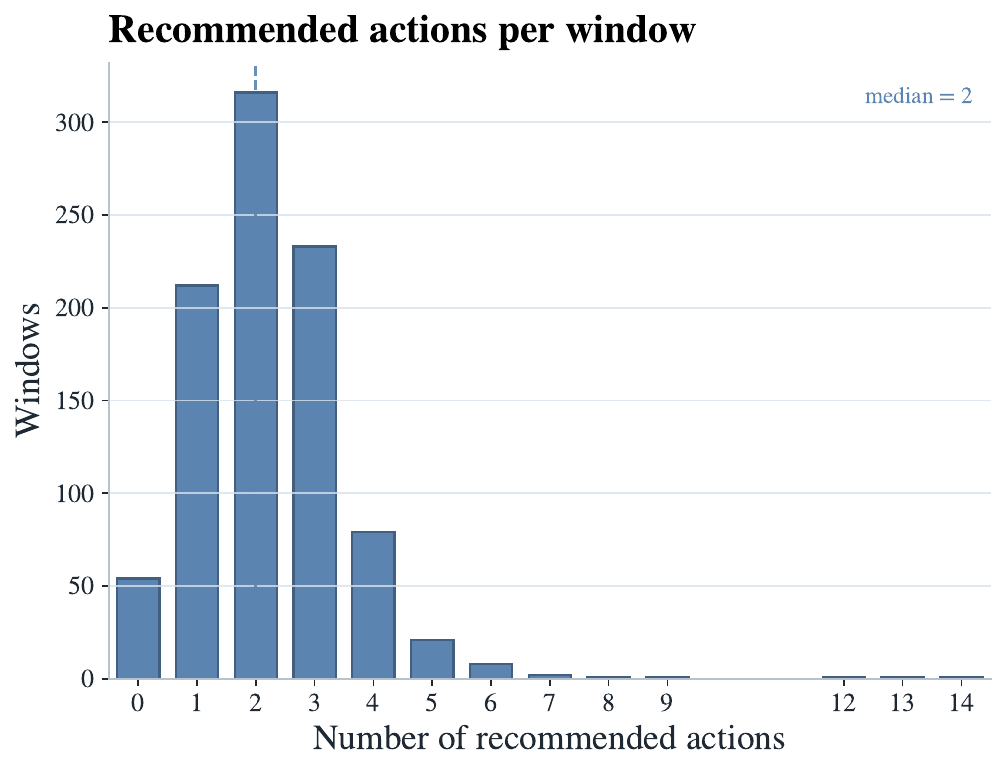}
    \end{subfigure}
    \hfill
    \begin{subfigure}[b]{0.3\textwidth}
        \includegraphics[width=\textwidth]{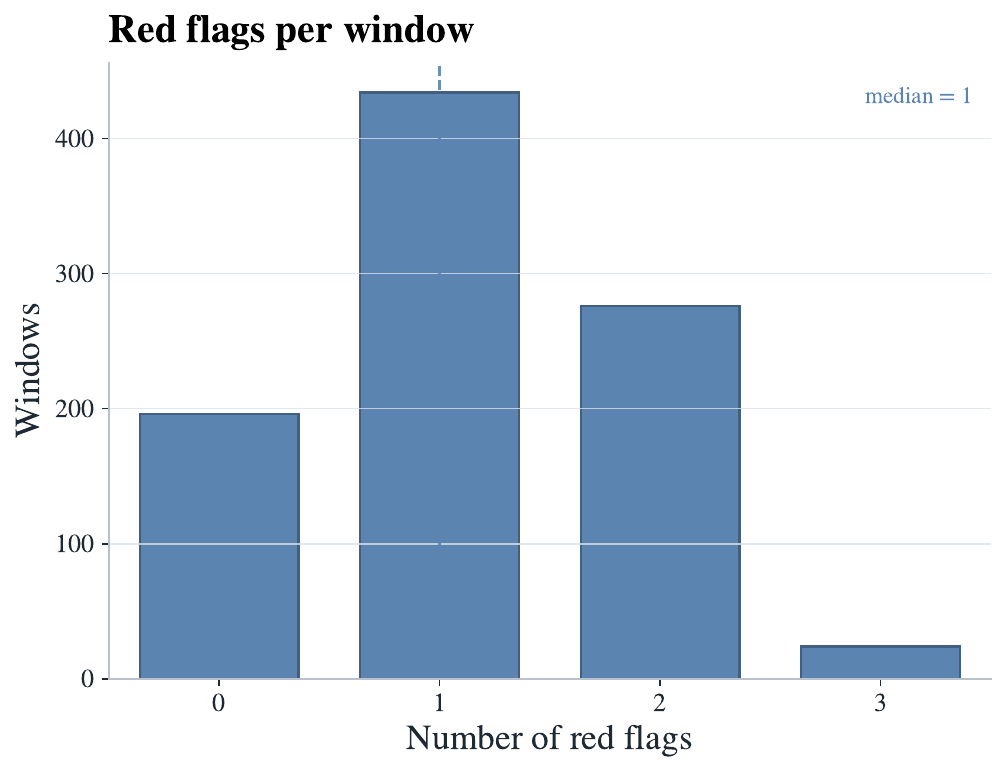}
    \end{subfigure}

    \caption{\emph{RealICU-Gold} statistics and label distribution for Patient Status, Active Problems, Recommended Action, and Red Flags. }
    \label{fig:realicu_gold_stats}
\end{figure}

\paragraph{\emph{RealICU-Gold} Label Statistics}
Figure~\ref{fig:realicu_gold_stats} summarizes the distributional properties of \textsc{RealICU-Gold}. The coverage histogram exhibits a long-tail distribution, with windows concentrated within the first 120 hours after ICU admission and a long right tail extending past 1{,}200 hours, yielding a median position of 74.8 hours. The \emph{Patient Status} distribution is dominated by Stable windows (63.0\%), followed by Deteriorating (22.4\%) and Improving (14.6\%). For the set-valued tasks, \emph{Acute Problems} is tightly concentrated around two concurrent problems per window, whereas \emph{Recommended Actions} exhibits a heavier-tailed distribution with a small number of windows reaching twelve or more concurrent recommendations, reflecting the variable cognitive load of ICU management. \emph{Red Flag Actions} remain rare by design, with a median of one per window and most windows containing zero or one event.

\begin{figure}[htbp]
    \centering
    \begin{subfigure}[b]{0.50\textwidth}
        \includegraphics[width=\textwidth]{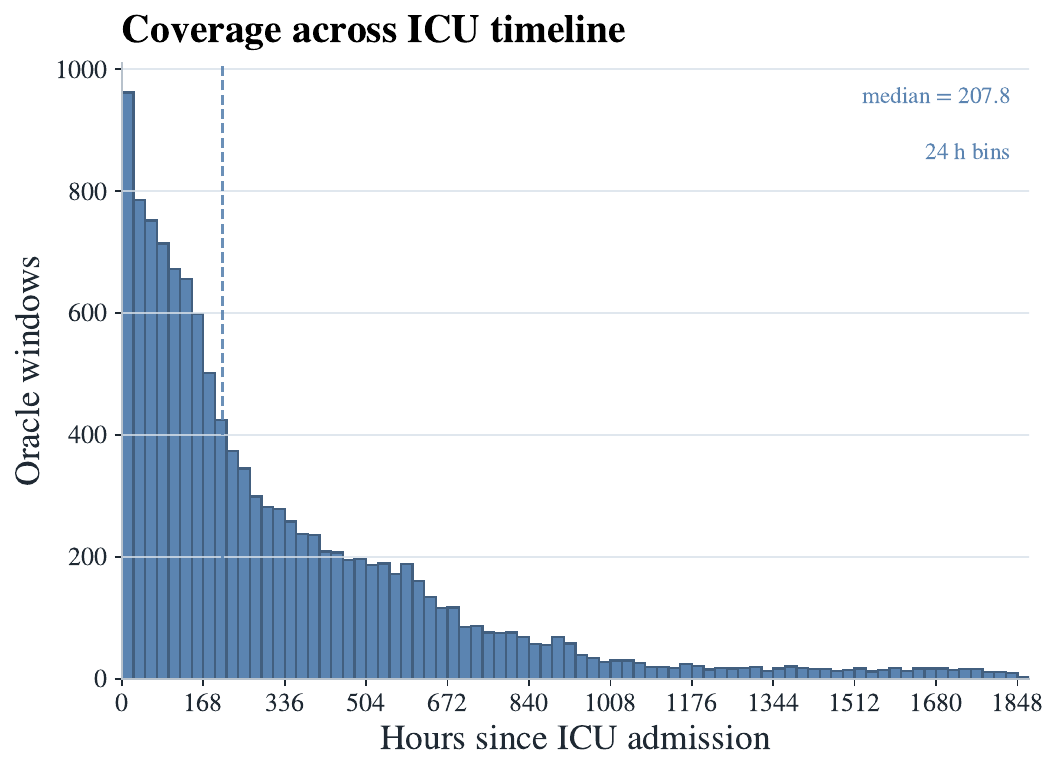}
    \end{subfigure}
    \hfill
    \begin{subfigure}[b]{0.35\textwidth}
        \includegraphics[width=\textwidth]{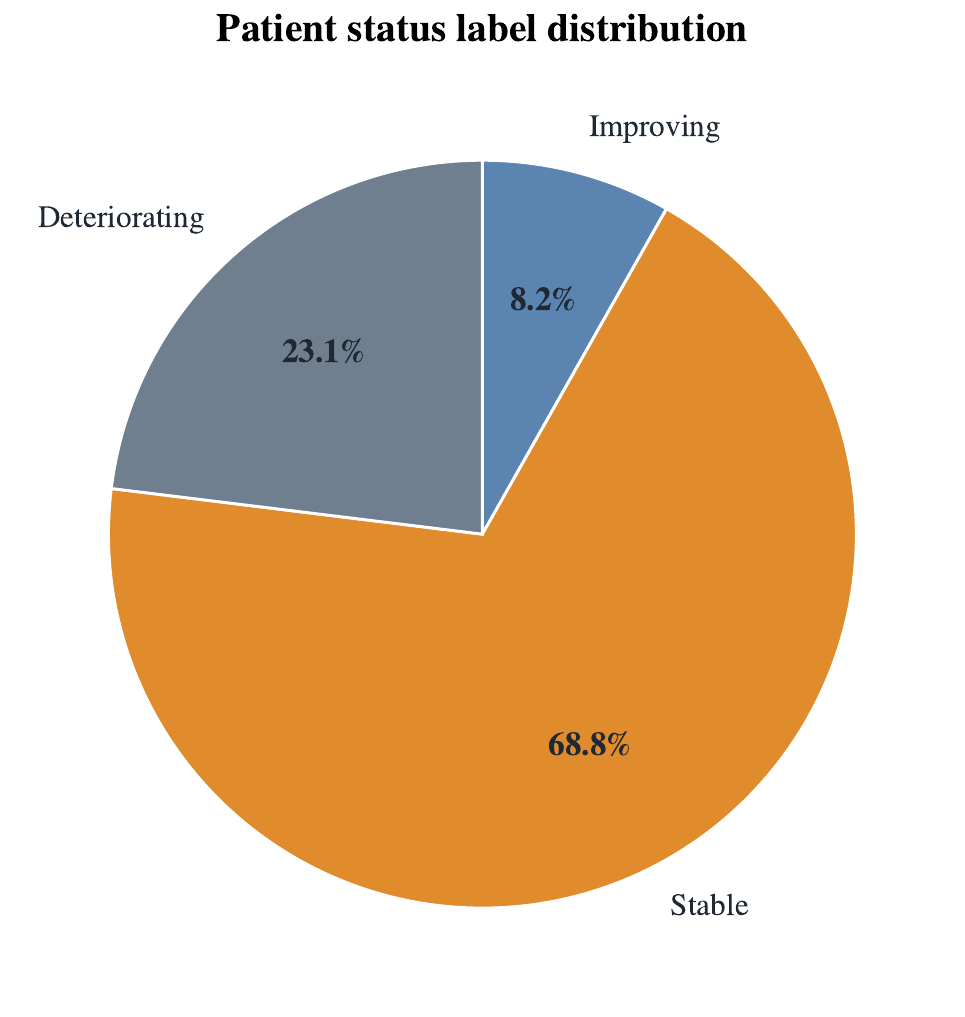}
    \end{subfigure}

    \vspace{0.5em}

    \begin{subfigure}[b]{0.3\textwidth}
        \includegraphics[width=\textwidth]{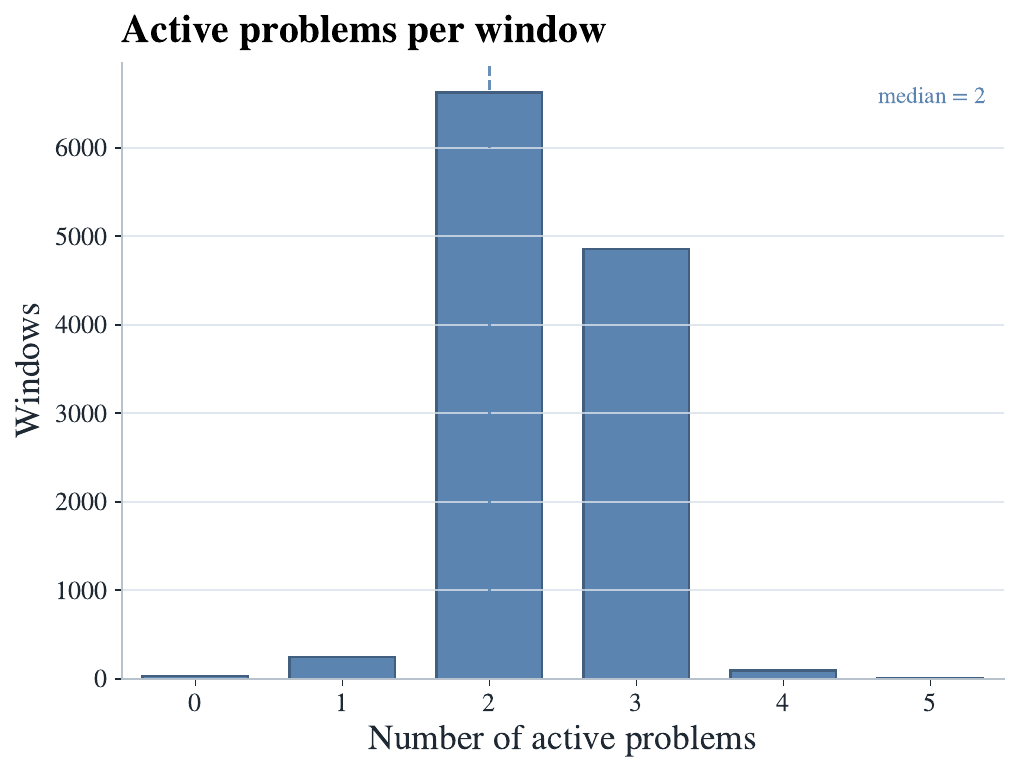}
    \end{subfigure}
    \hfill
    \begin{subfigure}[b]{0.3\textwidth}
        \includegraphics[width=\textwidth]{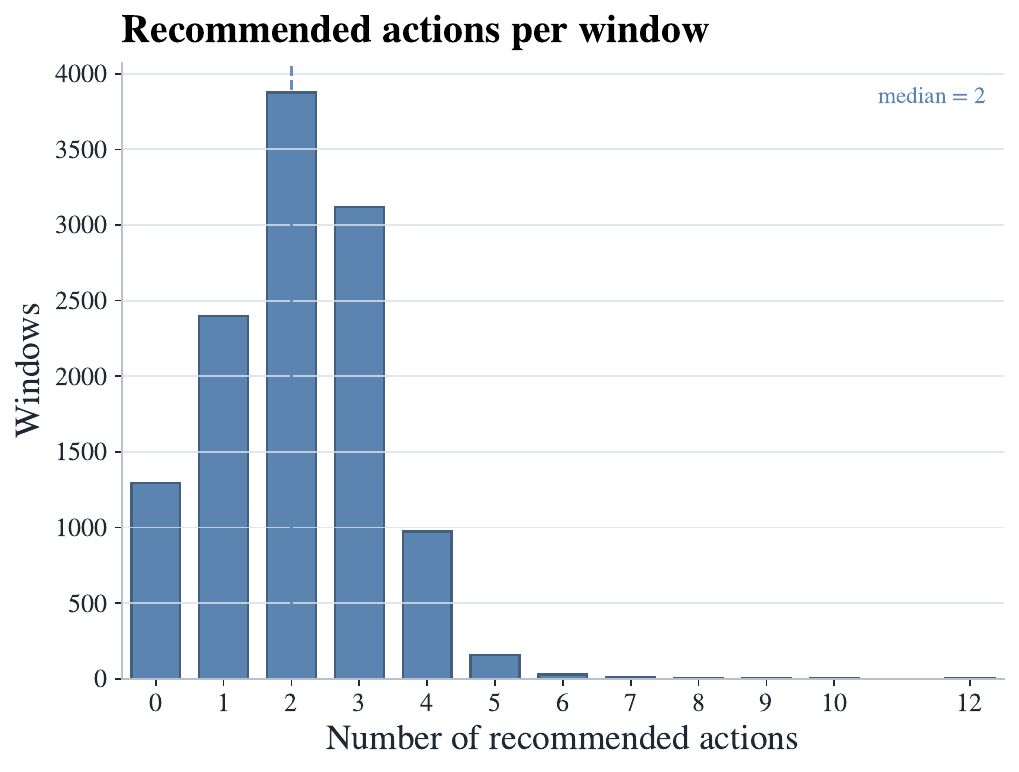}
    \end{subfigure}
    \hfill
    \begin{subfigure}[b]{0.3\textwidth}
        \includegraphics[width=\textwidth]{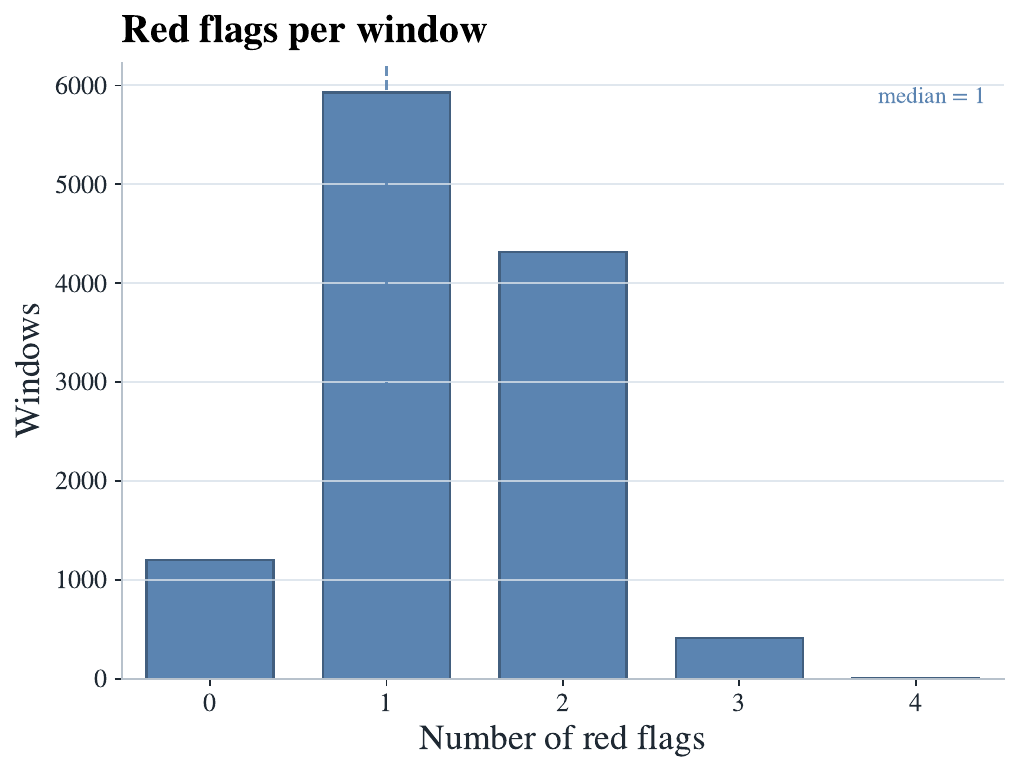}
    \end{subfigure}

    \caption{\emph{RealICU-Scale} statistics and label distribution for Patient Status, Active Problems, Recommended Action, and Red Flags. }
    \label{fig:realicu_scale_stats}
\end{figure}

\paragraph{\emph{RealICU-Scale} Label Statistics}
Figure~\ref{fig:realicu_scale_stats} summarizes the distributional properties of \emph{RealICU-Scale}. The coverage histogram exhibits a long-tail distribution, with windows concentrated within the first 336 hours after ICU admission and a long right tail extending past 1{,}800 hours, yielding a median position of 207.8 hours. The \emph{Patient Status} distribution is dominated by Stable windows (68.8\%), followed by Deteriorating (23.1\%) and Improving (8.2\%). For the set-valued tasks, \emph{Acute Problems} is tightly concentrated around two to three concurrent problems per window, whereas \emph{Recommended Actions} exhibits a heavier-tailed distribution with a small number of windows reaching ten or more concurrent recommendations, reflecting the variable cognitive load of ICU management. \emph{Red Flag Actions} has a median of one per window and most windows containing zero or one event.

\clearpage

\subsection{\emph{RealICU-Gold} Cross Validation}
\label{sec:app_cross_validation}
\emph{RealICU-Gold} contains 930 windows in total. For each window, we invite at least two out of five senior physicians for annotation. And we run a cross-validation check after annotation to maintain the golden-standard labels. In Table~\ref{tab:gold-stats}, we report the detailed number of each labels before and after cross-validation. Only labels with agreements are kept into \emph{RealICU-Gold}. Note that \emph{Active Problems},  \emph{Action Recomm.}, and \emph{Red Flags} are stored as sets with multiple labels per window. 

\begin{table}[h]
\centering
\small
\setlength{\tabcolsep}{6pt}
\caption{Label-wise statistics of \textsc{RealICU-Gold} after cross-validation filtering. }
\label{tab:gold-stats}
\begin{tabular}{lccc}
\toprule
\textbf{Task} & \textbf{N labels raw} & \textbf{N labels kept} & \textbf{Keep rate} \\
\midrule
\emph{Patient Status}        & 930    & 921    & 99.0\% \\
\emph{Active Problems}       & 2{,}170 & 2{,}066 & 95.2\% \\
\emph{Action Recomm.} & 2{,}328 & 2{,}198 & 94.4\% \\
\emph{Red Flags}             & 1{,}220 & 1{,}058 & 86.7\% \\
\bottomrule
\end{tabular}
\end{table}

\subsection{Dataset Pre-processing}
\label{sec:data-preproc}

To obtain our underlying base dataset of trajectories that cover ICU stays as well as their preceding patient journey, we merge MIMIC-IV~\cite{johnson2023mimic}, MIMIC-ED~\cite{johnson2023mimic}, MIMIC-Note~\cite{johnson2023mimic}, MIMIC-IV-ECHO~\cite{johnson2023mimic}, MIMIC-IV-ECG and MIMIC-CXR~\cite{johnson2019mimic}.

By this, we include not only patient meta data such as demographics, insurance, etc., but a diverse holistic timeline of medication, online medical records, vital measurements, X-ray, electro- and echocardiograms, procedures, diagnosis, lab results, text reports, and transfers.  We also include triaging data, subject to availability. 
From MIMIC-Note, we use the entire contents of the discharge summaries and the findings sections from radiology reports. In total, our resulting base dataset comprises 73,181 ICU stays from 50,920 patients. 

We arrange all charted information and measurements along a time axis together with patient age and time delta to the beginning of the specific ICU stay and sort them temporally ascending. Full duplicates are eliminated. Encoded categorical information from established ontologies and coding systems, e.g. for diagnosis (ICD) or medication (GSN), are resolved to their full-text descriptions. Text data is cleaned according to a permissive policy, only adjusting e.g. consecutive whitespace characters and unambiguous processing artifacts. Numerical data is also represented textually together with the respective unit of measurement and description. While we directly include all textually representable information and numeric measurements, we limit the integration of imaging and waveforms to their metadata, leaving the utilization of the X-ray, ECG, and ECHO contents to future work. 
We ensure that patient data is not leaked across our dataset splits. 

Further, we account for inaccurate charting and limitations of raw data collection by conservatively establishing an adversarial tolerance of 24h for key events such as discharge. In case of multiple records for the same event with different precision (e.g. death), usually originating from different tables in the raw dataset, we default to the most fine-grain timestamp. 

%% file: chapters/appendix/icu_evo.tex
\newpage
\section{Memory-Augmented Agents for Clinical Decision Support}
\label{sec:app_icu_evo}
\label{app:icu-evo}

We position ICU-Evo as an instance of the broader class of
memory-augmented language agents. In the following, we discuss a generic
formulation of the class, several specific instantiations
from recent work, and the design choices that motivate ICU-Evo.

\subsection{Formulation}
\label{sec:app_icu_evo_formulation}

A memory-augmented agent processes a stream of inputs
$\{x_1, x_2, \dots, x_T\}$ while maintaining an evolving memory state.
At step $t$, the update and decision rules take the generic form
\begin{equation}
M_t = \mathcal{U}(M_{t-1},\, x_t),
\qquad
y_t^{(k)} = f^{(k)}(M_t),
\label{eq:generic_memory_agent}
\end{equation}
where $M_t$ is the memory state, $\mathcal{U}$ is an update operator
that integrates the latest input into memory, and $f^{(k)}$ is a
task-specific decision function realized as a prompted call of the
underlying language model. Different memory systems differ primarily
in the structure of $M_t$ and in the choice of $\mathcal{U}$, and the
structural choices that define a memory system reduce to three
questions. What \emph{types} of content does $M_t$ contain, at what
\emph{temporal scale} is each type maintained, and under what
\emph{update policy} does each type evolve?

\subsection{Instantiations}
\label{sec:app_icu_evo_instantiations}

We describe three instantiations of Eq.~\ref{eq:generic_memory_agent}
in which $M_t$ takes increasingly heterogeneous forms.

\paragraph{Compressive Stream Memory.}
AgentFold~\citep{ye2025agentfold} sets $M_t$ as an ordered sequence of
summary blocks together with a high-fidelity record of the latest
interaction. The update operator $\mathcal{U}$ is a learned folding
policy that, at each step, either condenses the latest interaction
into a fine-grained block or consolidates a contiguous span of prior
blocks into a single coarse-grained block. This instantiation supports
streaming inputs and adaptive scale, while committing all memory
content to a single representational type (textual summary) under a
single update rule (replacement by summarization).

\paragraph{Cross-Task Experience Memory.}
Evo-Memory~\citep{wei2025evo} sets $M_t$ as an unordered set of
prior task experiences, each encoded as a structured tuple
$(x_i, \hat{y}_i, f_i)$, where $f_i$ is a feedback signal. The update
operator $\mathcal{U}$ is append-with-pruning, and a separate
refine action lets the agent reorganize or discard memory
entries during decision-making. This instantiation targets cross-task
transfer rather than within-task dynamics, and treats each task as the
atomic unit of memory.

\paragraph{Linked Note Memory.}
A-Mem~\citep{xu2025mem} sets $M_t$ as a collection of atomic notes,
where each note is a tuple of raw content, timestamp, LLM-generated
keywords, tags, and contextual description, a dense embedding, and a
set of links to other notes. The update operator $\mathcal{U}$ is
realized in two LLM-driven steps. On arrival of a new note, top-$k$
retrieval over the embedding space surfaces candidate neighbors, and
an LLM decides which neighbors deserve a semantic link. The same
neighbors are then re-examined, and the LLM may rewrite the
contextual description, keywords, or tags of any neighbor in light of
the new note. This instantiation supports streaming inputs and
introduces evolution of prior entries, while committing all memory
content to a single note schema under a single LLM-driven update rule.

\begin{table}[ht]
\centering
\small
\caption{ICU-Evo's memory components and the corresponding agent update operator.}
\label{tab:memory}
\begin{tabular}{@{}llll@{}}
\toprule
\textbf{Component} & \textbf{Definition} & \textbf{Updated by} \\
\midrule
$M^{\mathrm{work}}$ & Recent raw observations at full resolution.  & Observation Agent \\
$M^{\mathrm{trend}}$ & Piecewise-constant segmentations of vitals and labs.  & Observation Agent \\
$M^{\mathrm{event}}$ & Append-only log of critical events.  & Assessment Agent \\
$M^{\mathrm{traj}}$ & Compressed episode-level narrative of the stay.  & Assessment Agent \\
$M^{\mathrm{insight}}$ & Patient-specific hypotheses with supporting and counter-evidence.  & Insight Agent \\
\bottomrule
\end{tabular}
\end{table}

\subsection{ICU-Evo as Heterogeneous Clinical Memory}
\label{app:icu-evo-system}

ICU-Evo sets $M_t$ as a tuple of five components,
\begin{equation}
M_t = \bigl\{
M_t^{\mathrm{work}},\;
M_t^{\mathrm{trend}},\;
M_t^{\mathrm{event}},\;
M_t^{\mathrm{traj}},\;
M_t^{\mathrm{insight}}
\bigr\},
\label{eq:icu_evo_memory}
\end{equation}
defined in Table~\ref{tab:memory}. Algorithm~\ref{alg:icu-evo}
formalizes the full inference loop and
the pipeline of three agents that realize the update operator $\mathcal{U}$ at
different temporal cadences over the shared memory state. At every
window $t$, the Observation Agent ingests the new measurements $x_t$
and updates $M_t^{\mathrm{work}}$ by per-window overwrite and
$M_t^{\mathrm{trend}}$ by piecewise aggregation. Every $k_a$ windows,
the Assessment Agent compresses the recent working and trend memory
into a trajectory summary $z_t$, appended to $M_t^{\mathrm{traj}}$ as
a multi-scale rollup, and detects newly emerging critical events
$\tilde{E}_t$, appended to $M_t^{\mathrm{event}}$ under severity
gating. Every $k_i$ windows, the Insight Agent proposes
patient-specific hypotheses, gathers supporting and counter-evidence
from $M_t^{\mathrm{event}}$, and the Orchestrator commits the
accepted hypotheses to $M_t^{\mathrm{insight}}$ via lifecycle
transitions. The Predictor then queries the consolidated memory state
to emit task-specific predictions $y_t^{(k)}$, decoupled from the
memory update cycle.

The five components form principled correspondences with prior
designs, recombined under a common formulation. $M_t^{\mathrm{work}}$ and $M_t^{\mathrm{traj}}$ mirrors the multi-scale summaries of AgentFold~\cite{ye2025agentfold}, the lifecycle-managed
update of $M_t^{\mathrm{insight}}$ mirrors both the rewriting of prior
notes in A-Mem~\cite{xu2025mem} and the refine action of Evo-Memory~\cite{wei2025evo}.

\begin{algorithm}[ht]
\caption{ICU-Evo Memory-Augmented Agent System.}
\label{alg:icu-evo}
\small
\begin{algorithmic}[1]
\Require LLM backbone $\mathcal{F}$; ICU stay $s$; window sequence $\{x_t\}_{t=1}^{T}$; static context $c$; agent periods $k_a, k_i$
\State Initialize memory $M_0$ \Comment{work, trend, event, traj, insight}
\For{each window $t = 1, \ldots, T$}
  \State $\bigl(M_t^{\mathrm{work}},\, M_t^{\mathrm{trend}}\bigr) \gets \mathrm{Observe}\bigl(M_{t-1}^{\mathrm{work}},\, M_{t-1}^{\mathrm{trend}},\, x_t\bigr)$
  \Comment{Observation Agent; every window}
  \If{$t \bmod k_a = 0$} \Comment{Assessment Agent fires every $k_a$ windows}
    \State $\bigl(z_t,\, \tilde{E}_t\bigr) \gets \mathcal{F}\bigl(M_{t-k_a:t}^{\mathrm{work}},\, M_{t-k_a:t}^{\mathrm{trend}}\bigr)$
    \State $M_t^{\mathrm{traj}} \gets M_{t-1}^{\mathrm{traj}} \cup \{z_t\}$; \quad $M_t^{\mathrm{event}} \gets M_{t-1}^{\mathrm{event}} \cup \tilde{E}_t$
  \EndIf
  \If{$t \bmod k_i = 0$} \Comment{Insight Agent fires every $k_i$ windows}
    \State $\Delta H \gets \mathcal{F}\bigl(M_{t-1}^{\mathrm{insight}},\, M_{t-k_i:t}^{\mathrm{event}}\bigr)$ \Comment{propose/update hypotheses}
    \For{each hypothesis $h \in \Delta H$}
      \State $\mathrm{state}(h) \gets \textit{accept}$ if $s(h) > r(h)$ else $\textit{reject}$
    \EndFor
    \State $M_t^{\mathrm{insight}} \gets \mathrm{Orchestrator}\bigl(M_{t-1}^{\mathrm{insight}},\, \Delta H\bigr)$
  \EndIf
  \For{each task $k$} \Comment{Predictor decoupled from memory update}
    \State $y_t^{(k)} \gets \mathcal{F}^{(k)}\bigl(M_t;\, c\bigr)$
  \EndFor
\EndFor
\State \Return predictions $\{y_t^{(k)}\}$ for evaluation against \emph{RealICU} labels
\end{algorithmic}
\end{algorithm}

\subsection{Discussion}
\label{sec:app_icu_evo_discussion}
The instantiations above demonstrate the flexibility of Eq.~\ref{eq:generic_memory_agent}, yet alternative combinations remain possible. The heterogeneous decomposition we adopt reflects that clinical reasoning under partial observability proceeds along multiple simultaneous modes. A homogeneous memory forces a single answer to three independent questions: at what temporal scale to retain content, at what fidelity, and under what update policy. AgentFold~\cite{ye2025agentfold} couples scale and fidelity under a uniform textual summary type, fitting neither append-only event logs nor lifecycle-managed hypotheses. A-Mem~\cite{xu2025mem} couples all three under a uniform note schema and LLM-driven evolution rule, providing no mechanism for the distinct update policies that event detection and hypothesis lifecycle management each require. Evo-Memory~\cite{wei2025evo} treats each task as the unit of experience, fitting cross-task transfer but leaving within-patient dynamics unaddressed. The heterogeneous memory structure of ICU-Evo in Eq.~\ref{eq:icu_evo_memory} offers a frameworks to align previous designs into real clinical reasoning over evolving patient states.

%% file: chapters/appendix/case_study.tex
\newpage
\section{Case Study}
\label{sec:app_case_study}
\label{app:casestudy}

\subsection{Failure Case: Recall Safety Tradeoff}
\label{sec:app_recall_safety}

Patient background: Age 55. Female, entering ICU with severe subarachnoid
hemorrhage. Course complicated by severe intracranial hypertension (ICP $>$40\,mmHg)
managed with mannitol, propofol, and right frontal EVD placement. Current 191.5--192.0 hours after ICU admission (day-8).

\tcbset{
  casestudybox/.style={
    colback=white,
    colframe=prompttitlebar,
    colbacktitle=prompttitlebar,
    coltitle=white,
    boxrule=0.8pt,
    fonttitle=\bfseries\small,
    fontupper=\footnotesize,
    arc=2pt,
    left=4pt, right=4pt, top=4pt, bottom=4pt,
    before skip=4pt, after skip=4pt,
  }
}

\begin{figure}[H]
\centering

\begin{tcolorbox}[casestudybox, title={\textbf{Current-window events}}]
\textbf{11:16}~~\gn{Impaired Tissue Perfusion NCP -- Therapeutic temperature management}\\
\textbf{11:16}~~\gn{Impaired Tissue Perfusion NCP -- Pressors}\\
\textbf{11:17}~~Infection NCP -- Monitor and assess respiratory secretions\\
\textbf{11:17}~~Impaired Tissue Perfusion NCP -- Monitor neuro assessment\\
\textbf{11:17}~~Impaired Tissue Perfusion NCP -- Monitor hemodynamic status q1h\\
\hspace*{1.6em}\ldots (63 further entries)
\end{tcolorbox}

\begin{tcolorbox}[casestudybox, title={\textbf{Memory snapshot}}]
\textbf{Working memory:} 68 events from the current window.

\smallskip
\textbf{Trend memory:} None at current window.

\smallskip
\textbf{Critical events:} 08:16 VITALS Richmond-RAS Scale=$-4.00$; 11:00 VITALS Intra Cranial Pressure 21.0\,mmHg; \ldots

\smallskip
\textbf{Trajectory:} This patient exhibits a highly coupled neuro-respiratory deterioration, demonstrating acute and severe hypoxemia simultaneously with sudden intracranial pressure spikes \ldots

\smallskip
\textbf{Insight \#6:} This patient exhibits extreme ICP volatility, demonstrating sudden, severe ICP spikes that break through maximal medical management, including deep sedation, neuromuscular blockade, and targeted temperature management. Expect a higher-than-average risk of sudden, refractory intracranial crises and secondary cerebral hypoperfusion despite seemingly stable baseline ICP, \rd{requiring aggressive, anticipatory hyperosmolar or surgical interventions.}
\end{tcolorbox}

\begin{tcolorbox}[casestudybox, title={\textbf{Gold annotation}}]
\textbf{Patient Status.} Deteriorating.

\smallskip
\textbf{Active Problems.} Refractory intracranial hypertension and impending herniation; \rd{worsening hypernatremia and hyperosmolality (Na 155\,mEq/L, osm 325\,mOsm/kg)}; ventilator-associated pneumonia.

\smallskip
\textbf{Recommended Actions.} \gn{Therapeutic temperature management}; monitor and assess respiratory secretions.

\smallskip
\textbf{Red Flags.} \rd{Routine hyperosmolar therapy (mannitol or hypertonic saline)} --- further routine administration is contraindicated and dangerous under current sodium and osmolality; emergency bolusing only as a last resort for acute herniation.
\end{tcolorbox}

\begin{tcolorbox}[casestudybox, title={\textbf{ICU-Evo predictions (recommended actions)}}]
\renewcommand{\arraystretch}{1.15}
\begin{tabularx}{\linewidth}{@{}r@{\hspace{4pt}}X@{\hspace{6pt}}l@{}}
1. & Urgent neurosurgery and neurocritical care consultation. & \textit{[unmatch]} \\
2. & \rd{Administer hyperosmolar therapy. Boluses of mannitol or 3\% hypertonic saline for sustained ICP $>$20--22\,mmHg.} & \textit{\rd{[red flag]}} \\
3. & \gn{Maintain CPP $>$70\,mmHg. Continuous norepinephrine to meet MAP goals.} & \textit{\gn{[match]}} \\
4. & Initiate goals-of-care discussion. & \textit{[unmatch]} \\
5. & Strict glycemic and electrolyte monitoring (q1--2h K, glucose). & \textit{[unmatch]} \\
\end{tabularx}
\end{tcolorbox}

\caption{Recall-safety tradeoff case study. ICU-Evo's stored insight \#6 prescribes ``aggressive, anticipatory hyperosmolar interventions,'' which propagates to prediction 2 --- flagged as contraindicated by the gold annotation under current Na/osm. The trend layer carries no sodium signal forward, so nothing counterweighs the stored frame.}
\label{fig:case_recall_safety}
\end{figure}

\clearpage
\subsection{Failure Case: Anchoring Bias}
\label{sec:app_anchoring_bias}

Patient background: Adult patient with Eisenmenger's physiology, severe right ventricular dysfunction, and chronic right-to-left shunting with polycythemia. Baseline SpO$_2$ 85--93\%. Prior course marked by refractory hypoxemia (SpO$_2$ nadir 77\% on FiO$_2$ 100\%) unresponsive to maximal high-flow oxygen and ongoing diuresis. Current 87.5--88.0 hours after ICU admission (day-4).

\begin{figure}[H]
\centering

\begin{tcolorbox}[casestudybox, title={\textbf{Current-window events}}]
\textbf{18:48}~~BODY\_INPUT Oral/Gastric Ingredient = 300\,ml\\
\textbf{18:48}~~BODY\_INPUT PO Intake = 300\,ml\\
\textbf{18:48}~~\gn{BODY\_INPUT Water = 300\,ml}\\
\textbf{18:48}~~\gn{VITALS Weight = 51.8\,kg}\\
\hspace*{1.6em}(4 events total; no vitals stream this window)
\end{tcolorbox}

\begin{tcolorbox}[casestudybox, title={\textbf{Memory snapshot}}]
\textbf{Working memory:} 4 events from the current window.

\smallskip
\textbf{Trend memory:} No vital signal this window (HR/RR/SpO$_2$/BP/MAP/Temp all empty).

\smallskip
\textbf{Critical events:} 02-03 13:00 VITALS SpO$_2$ 77\% on FiO$_2$ 100\%; 02-04 06:00 VITALS O$_2$ Flow 35\,L/min; 02-04 22:00 BODY\_OUTPUT Void 700\,ml; \ldots

\smallskip
\textbf{Trajectory:} This patient demonstrates a coupled refractory-hypoxemia and preload-dependence pattern, where standard oxygen escalation fails to correct shunt-driven desaturation and successful volume offloading precipitates hemodynamic collapse \ldots

\smallskip
\textbf{Insight \#2:} This patient demonstrates refractory hypoxemia that is unresponsive to maximal high-flow oxygen therapy and ongoing diuresis. Expect below-average improvement in oxygenation indices from standard volume offloading, \rd{likely due to underlying pulmonary hypertension and fixed V/Q mismatch or shunting.}
\end{tcolorbox}

\begin{tcolorbox}[casestudybox, title={\textbf{Gold annotation}}]
\textbf{Patient Status.} Stable.

\smallskip
\textbf{Active Problems.} Hypotension and reduced cardiac output (preload-sensitive); volume overload risk under chronic right-heart strain.

\smallskip
\textbf{Recommended Actions.} \gn{BODY\_INPUT Water (oral intake)}; \gn{VITALS Weight (daily tracking)}.

\smallskip
\textbf{Red Flags.} \rd{Titrating supplemental oxygen to SpO$_2$ $>$ 95\%} --- baseline SpO$_2$ is 85--93\% from chronic right-to-left shunting and polycythemia; oxygen should be administered only if SpO$_2$ drops below 85\% or the patient becomes symptomatic.
\end{tcolorbox}

\begin{tcolorbox}[casestudybox, title={\textbf{ICU-Evo predictions (recommended actions)}}]
\renewcommand{\arraystretch}{1.15}
\begin{tabularx}{\linewidth}{@{}r@{\hspace{4pt}}X@{\hspace{6pt}}l@{}}
1. & Hold or reduce diuretic therapy. Suspend or decrease loop diuretics and reassess volume status before further diuresis. & \textit{[unmatch]} \\
2. & Titrate norepinephrine to maintain MAP $>$ 65\,mmHg, weaning cautiously if hemodynamics remain stable. & \textit{[unmatch]} \\
3. & Monitor serum potassium via basic metabolic panel or venous blood gas. & \textit{[unmatch]} \\
4. & \rd{Maintain targeted oxygenation. Continue 4\,L/min nasal cannula to target SpO$_2$ 88--92\%, avoiding over-oxygenation.} & \textit{\rd{[red flag]}} \\
\end{tabularx}
\end{tcolorbox}

\caption{Premature-anchoring case study. The window contains four events --- oral water and a daily weight --- and the gold status is stable. ICU-Evo's stored insight \#2 carries forward the prior cardiopulmonary story of refractory hypoxemia, propagating to prediction 4, which prescribes an active oxygen target (88--92\%) that the gold annotation marks as contraindicated under this patient's Eisenmenger baseline. The trend layer is empty for the current window, so nothing pulls the model back to the simpler window-grounded interpretation.}
\label{fig:case_premature_anchoring}
\end{figure}

\clearpage
\subsection{Memory Snapshot}
\label{sec:memory_snapshot}

\begin{figure}[H]
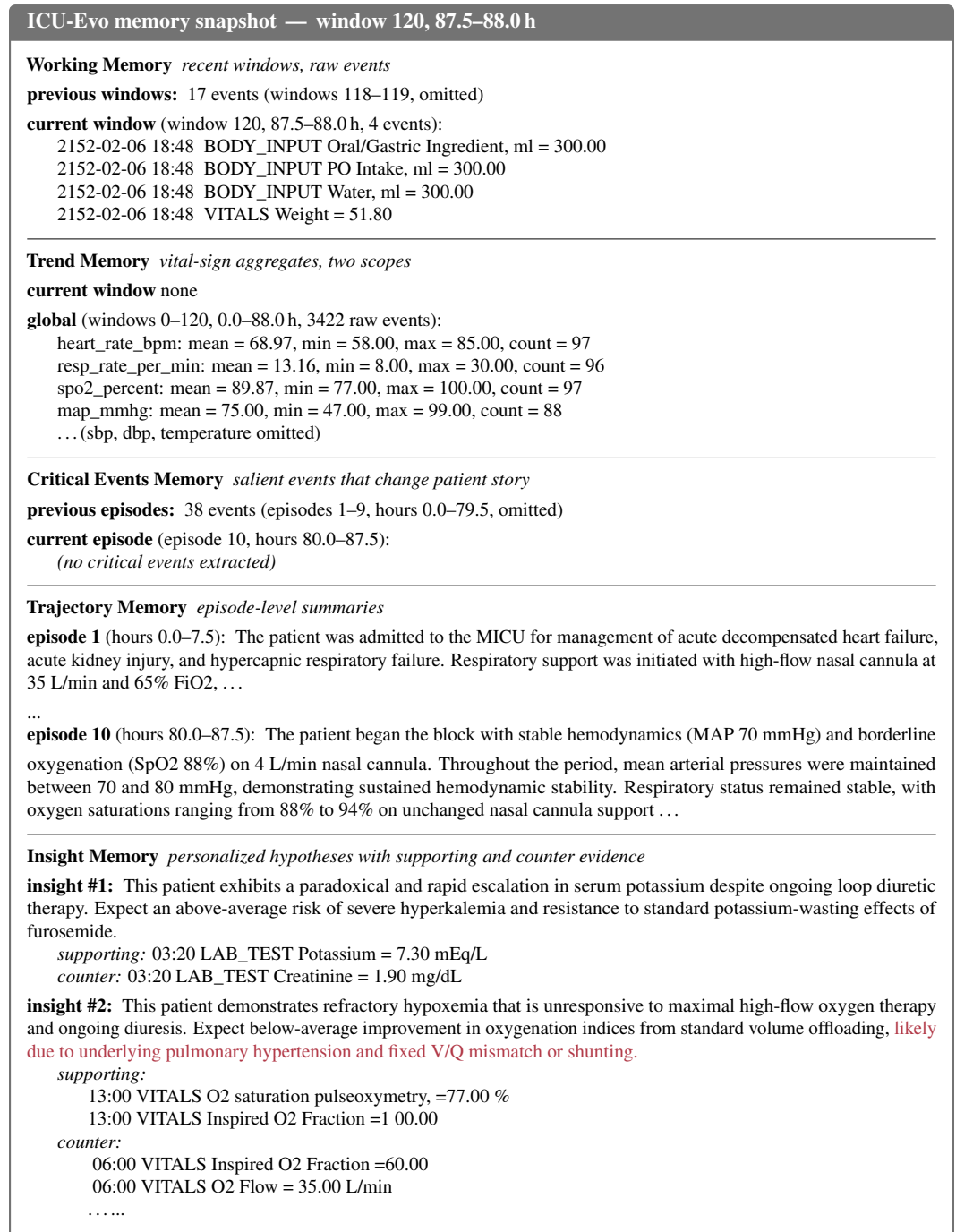

\centering

\begin{tcolorbox}[casestudybox, title={\textbf{ICU-Evo memory snapshot \,---\, window 120, 87.5--88.0\,h}}]

\textbf{Working Memory}~~\textit{recent windows, raw events}

\smallskip
\textbf{previous windows:}~~17 events (windows 118--119, omitted)

\smallskip
\textbf{current window} (window 120, 87.5--88.0\,h, 4 events): \\
\hspace*{1.6em}2152-02-06 18:48~~BODY\_INPUT Oral/Gastric Ingredient, ml = 300.00\\
\hspace*{1.6em}2152-02-06 18:48~~BODY\_INPUT PO Intake, ml = 300.00\\
\hspace*{1.6em}2152-02-06 18:48~~BODY\_INPUT Water, ml = 300.00\\
\hspace*{1.6em}2152-02-06 18:48~~VITALS Weight = 51.80

\medskip
\hrule height 0.3pt
\medskip

\textbf{Trend Memory}~~\textit{vital-sign aggregates, two scopes}

\smallskip
\textbf{current window} none

\smallskip
\textbf{global} (windows 0--120, 0.0--88.0\,h, 3422 raw events):\\
\hspace*{1.6em}heart\_rate\_bpm: mean = 68.97, min = 58.00, max = 85.00, count = 97\\
\hspace*{1.6em}resp\_rate\_per\_min: mean = 13.16, min = 8.00, max = 30.00, count = 96\\
\hspace*{1.6em}spo2\_percent: mean = 89.87, min = 77.00, max = 100.00, count = 97\\
\hspace*{1.6em}map\_mmhg: mean = 75.00, min = 47.00, max = 99.00, count = 88\\
\hspace*{1.6em}\ldots (sbp, dbp, temperature omitted)

\medskip
\hrule height 0.3pt
\medskip

\textbf{Critical Events Memory}~~\textit{salient events that change patient story}

\smallskip
\textbf{previous episodes:}~~38 events (episodes 1--9, hours 0.0--79.5, omitted)

\smallskip
\textbf{current episode} (episode 10, hours 80.0--87.5):\\
\hspace*{1.6em}\textit{(no critical events extracted)}

\medskip
\hrule height 0.3pt
\medskip

\textbf{Trajectory Memory}~~\textit{episode-level summaries}

\smallskip
\textbf{episode 1} (hours 0.0--7.5):~~The patient was admitted to the MICU for management of acute decompensated heart failure, acute kidney injury, and hypercapnic respiratory failure. Respiratory support was initiated with high-flow nasal cannula at 35 L/min and 65\% FiO2, \ldots

\smallskip
... \\
\smallskip
\textbf{episode 10} (hours 80.0--87.5):~~The patient began the block with stable hemodynamics (MAP 70 mmHg) and borderline oxygenation (SpO2 88\%) on 4 L/min nasal cannula. Throughout the period, mean arterial pressures were maintained between 70 and 80 mmHg, demonstrating sustained hemodynamic stability. Respiratory status remained stable, with oxygen saturations ranging from 88\% to 94\% on unchanged nasal cannula support \ldots

\medskip
\hrule height 0.3pt
\medskip

\textbf{Insight Memory}~~\textit{personalized hypotheses with supporting and counter evidence}

\smallskip
\textbf{insight \#1:}~~This patient exhibits a paradoxical and rapid escalation in serum potassium despite ongoing loop diuretic therapy. Expect an above-average risk of severe hyperkalemia and resistance to standard potassium-wasting effects of furosemide.\\
\hspace*{1.6em}\textit{supporting:} 03:20 LAB\_TEST Potassium = 7.30 mEq/L \\
\hspace*{1.6em}\textit{counter:} 03:20 LAB\_TEST Creatinine = 1.90 mg/dL 

\smallskip
\textbf{insight \#2:}~~This patient demonstrates refractory hypoxemia that is unresponsive to maximal high-flow oxygen therapy and ongoing diuresis. Expect below-average improvement in oxygenation indices from standard volume offloading, \rd{likely due to underlying pulmonary hypertension and fixed V/Q mismatch or shunting.}\\
\hspace*{1.6em}\textit{supporting:}\\
\hspace*{1.6em}\hspace*{1.6em}13:00 VITALS O2 saturation pulseoxymetry, =77.00 \% \\
\hspace*{1.6em}\hspace*{1.6em}13:00 VITALS Inspired O2 Fraction =1 00.00\\
\hspace*{1.6em}\textit{counter:}\\
\hspace*{1.6em}\hspace*{1.6em} 06:00 VITALS Inspired O2 Fraction =60.00\\
\hspace*{1.6em}\hspace*{1.6em} 06:00 VITALS O2 Flow = 35.00 L/min\\
\hspace*{1.6em}\hspace*{1.6em}\ldots
\smallskip
...

\end{tcolorbox}

\caption{ICU-Evo memory snapshot at 87.5--88.0 hours after admission. The five layers of memory together constitute the full state available to the prediction modules at this window, including working memory, trend, critical events, trajectory, and patient-specific insights. Red highlights mark the thread most relevant to the case study in Figure~\ref{fig:case_premature_anchoring}.}
\label{fig:memory_snapshot}

\end{figure}

%% file: chapters/appendix/prompts.tex
\newpage
\section{Prompts}
\label{sec:app_prompts}
\label{app:prompts}
\begingroup
\setlength{\parindent}{0pt}

\subsection{Oracle Prompt}
\label{sec:app_oracle_prompt}
\begin{promptbox}{Oracle Prompt}
You are Oracle, a clinical AI evaluator with hindsight access to a patient's full ICU trajectory.
Your task is to evaluate a specific local observation window {window_time} across two parts.

=== PART 1: PATIENT ASSESSMENT ===

[1A. CURRENT STATUS]

Assess the patient's clinical direction at this window by reasoning across four domains: Hemodynamics, Respiratory, Renal/Metabolic, Neurology

Synthesize your domain reasoning into a single overall status label:
- improving: indicators trending toward recovery relative to the provided context
- stable: no meaningful change in either direction
- deteriorating: indicators trending toward worsening relative to the provided context
- insufficient_data: available information is not sufficient to make a reliable judgment (e.g., sparse events, conflicting signals)

Important nuances:
- Do NOT conflate outcome with care quality. A patient may be deteriorating despite excellent care.
- A patient may be labeled stable or improving even if they eventually die, if the trajectory at this window genuinely reflects that direction.

[1B. ACTIVE PROBLEMS]

Using the full trajectory and current window, identify active clinical problems or emerging risks this patient faces going forward from this window.
- Only include risks that are real and imminent or already developing --- not distant or hypothetical
- Each risk must be tied to specific trajectory evidence
- An empty list is expected and acceptable when no urgent risks are present

=== PART 2: ACTION REVIEW ===

[2A. ACTION EVALUATION]

Evaluate each clinical action taken during this window. For each action, integrate two perspectives into a single judgment:
- Guideline alignment: does this action follow established ICU guidelines (e.g., Surviving Sepsis Campaign, ARDSNet, PADIS, AHA/ACC where relevant)?
- Contextual appropriateness: given this patient's specific condition, trajectory, comorbidities, and the eventual outcome known to you, was this action appropriate?

Assign one overall label:
- best_practice: action is both guideline-aligned and well-suited to this patient's specific situation
- acceptable: action is reasonable given the context, even if not optimal or if guidelines are ambiguous
- potentially_harmful: action poses real risk of harm to this patient, whether due to guideline violation, patient-specific contraindication, or both
- insufficient_data: not enough context to evaluate this action reliably

Use your hindsight knowledge to inform the judgment, but be fair: judge the action against what the context reveals, not against impossible foresight. If an action was reasonable at the time but later context revealed a missed diagnosis, note this nuance explicitly in the rationale.

The action can be identified by its code, which includes and are not limited to: DRUG_START, DRUG_STOP, DRUG_PRESCRIPTION, BODY_INPUT, TRANSFER, LAB_TEST, DIAGNOSIS.

[2B. RED FLAGS]

Using the full trajectory and current window, identify any actions that should be strictly avoided for this specific patient going forward.
- Only flag actions that a reasonable clinician might consider but would be harmful for this specific patient --- do not list generic contraindications unless directly applicable here
- Each flag must be justified by patient-level evidence (comorbidities, trajectory events, organ function, known sensitivities)
- An empty list is expected and acceptable when no red flags are present

=== PATIENT ICU CONTEXT WINDOW ===

{patient_icu_trajectory}

Now, evaluate the CURRENT OBSERVATION WINDOW according to the instructions above.
\end{promptbox}

\subsection{Agent Prompt}
\label{sec:app_agent_prompt}

\begin{promptbox}{Assessment Agent Prompt}
You are an ICU assessment agent. You compress {k} consecutive 30-minute windows
({duration} hours total) into one episode summary and identify the block's
critical events.

=== INPUT ===

[PATIENT METADATA] (read-only context)
{patient_metadata}

[PRIOR EPISODE SUMMARY] (read-only context)
Used only to distinguish new from continuing findings and to calibrate baseline.
Do not cite from it.
{prior_episode_summary_text}

[EPISODE TIME RANGE]
{episode_start_time} to {episode_end_time}

[WINDOWED ICU DATA]
You receive:
- Raw events for each window, formatted as `[<event_id>] <time> <event_name> <payload>`
- One grouped `selected vital trends` section summarizing tracked vitals across the whole block
- In the trend section, only windows with at least one value for that vital are shown,
  plus an overall summary across the block

{episode_input}

=== TASK ===

Produce (1) an episode summary and (2) a list of critical events.

[TASK 1: EPISODE SUMMARY] (3--6 sentences)

Narrate the block's clinical trajectory: where the patient started the block, what
meaningfully happened, where they ended, and what remains unresolved. Focus on
direction of travel and inflection points, not a window-by-window recap.
- Use specific numeric values at inflection points; avoid adjective-only descriptions
  ("rising", "unstable") without numbers.
- Include sustained abnormal states only when their persistence is the point
  (e.g., tachycardia held across the block despite intervention).
- When the prior summary establishes a condition or intervention, frame current
  findings as continuation, escalation, or resolution, not new onset.
- Stay descriptive and temporal. Do not assign diagnoses, syndromes, or mechanisms
  that are not explicitly in the input.
- Refer to events by clinical name in prose. Event IDs appear only in
  `supporting_event_ids`.

[TASK 2: CRITICAL EVENTS]

A critical event is a high-SNR inflection point in the patient's ICU trajectory:
a moment that materially changes the clinical story. Reading only the critical
events, a clinician should be able to reconstruct the shape of the block.

Critical events typically fall into one of these categories:
- New or worsening organ dysfunction (respiratory, cardiovascular, renal, hepatic,
  neurological).
- Resolution or meaningful improvement of existing organ dysfunction.
- Initiation of a major intervention (intubation, vasopressor start, dialysis,
  transfusion for active bleed, emergency procedure).
- Diagnosis scores, such as GCS, RASS, or SOFA scores.
- Significant escalation or de-escalation of care reflecting a change in trajectory.

An event qualifies only if it is clinically meaningful on its own, changes how
subsequent data should be read, and is corroborated by surrounding trend or events
rather than an isolated outlier.
Exclude routine readings, scheduled medications without clinical context, or noisy
measurements.
Err toward under-listing. If no event meets the bar, return an empty list.

\end{promptbox}

\begin{promptbox}{Insight Agent}
You are a clinical insight agent. Your job is to identify how THIS patient deviates
from the population-average ICU patient in similar circumstances, specifically in how
they respond to illness and interventions, or how their physiology is trending
relative to what the current illness and treatment would predict.

You are generating patient-specific response profiles and trajectory deviations that
would change how future data should be interpreted or how future decisions should be
weighted.

You will be given:
- The patient's existing hypothesis bank (all hypotheses, active and retired)
- Patient metadata with compressed pre-ICU history, for background context only
- The latest episode:
  - Clinical trajectory summary
  - Critical events, formatted as `<event_id> <time> <event_name> <payload>`
  - Vital trend statistics

=== INPUT ===

[EXISTING HYPOTHESES]
{hypothesis_bank}

[PATIENT METADATA]
{patient_metadata}

[LATEST EPISODE SUMMARY]
{episode_summary}

[LATEST EPISODE CRITICAL EVENTS]
{critical_events}

[LATEST EPISODE VITAL TRENDS]
{vital_trends}

=== TASK ===

[TASK 1: EVIDENCE FOR EXISTING HYPOTHESES]

This is a matching task. Scan the episode for evidence bearing on each active
hypothesis. Do not infer beyond what is stated.
For each hypothesis where the episode provides relevant signal, report:
- `supporting_evidence`: event IDs or `vital_trend` that reinforce it
- `counter_evidence`: event IDs or `vital_trend` that weaken it
Updates must cite at least one piece of evidence. No citation, no update.
Skip hypotheses with no relevant signal.

[TASK 2: NEW HYPOTHESIS GENERATION]

This task requires reasoning. Before proposing a new hypothesis, mentally construct
the population-average trajectory for a patient with this illness receiving these
interventions, then compare to what you observe. Only flag a new hypothesis where
this patient's pattern meaningfully departs from that reference.

A valid new hypothesis must:
- Describe an individualized response profile or trajectory deviation (not a
  diagnosis, not an event restatement)
- Be grounded in evidence
- Include an expected deviation from population-average response (e.g.,
  "below-average," "slower than typical")
- Be clinically actionable or prognostically relevant

Over-generation is worse than under-generation. Generate at most 2 new hypotheses;
if more candidates exist, report only the strongest.
Do not flag obvious, trivial, or diagnosis-shaped claims. Empty list is acceptable
and often correct.

[GROUNDING RULES]
- Cite only event IDs that appear in the critical events block, or the token
  `vital_trend` for claims grounded in trend statistics.
- Do not invent events, values, or clinical facts not present in the provided inputs.

[EXAMPLES OF VALID NEW HYPOTHESES]

Example A --- individualized response profile:
"This patient shows diminished hemodynamic response to fluid resuscitation.
Expect below-average MAP rise to standard fluid boluses."

Example B --- trajectory deviation:
"This patient's respiratory recovery is progressing slower than typical for their
current ventilator settings and sedation level. Expect below-average improvement
in oxygenation indices over the next shift."

\end{promptbox}

\subsubsection{Predictor Prompt}

\begin{promptbox}{Shared Prompt}
You are a clinical decision support AI. Assess the patient's overall clinical
status for the current ICU window.

=== INPUT ===

You will receive one of:

(A) Raw ICU events in chronological order

(B) A structured Memory object with the following layers:
- patient_metadata
- working_memory         : raw events from recent windows --- current local status
- trend_memory           : per-window vital trend statistics
- critical_events_memory : episode-level critical events
- trajectory_memory      : episode summaries
- insights               : patient-specific deviations from typical ICU trajectories

\end{promptbox}

\begin{promptbox}{Patient Status Predictor}

=== INSTRUCTIONS ===

Assess the patient's clinical direction at this window by reasoning across four
domains:
- Hemodynamics   : heart rate, MAP, lactate, perfusion, vasopressor requirements
- Respiratory    : SpO2, PaO2/FiO2, respiratory rate, ventilator settings
- Renal/Metabolic: creatinine, urine output, electrolytes, acid-base status
- Neurology      : GCS, mental status, RASS sedation score

Synthesize your domain reasoning into a single overall status label, weighted by
the relative clinical importance of each domain for this specific patient:
- improving         : indicators trending toward stability or recovery relative
                      to the provided context
- stable            : no meaningful change in either direction
- deteriorating     : indicators trending toward worsening or decompensation
                      relative to the provided context
- insufficient_data : available information is not sufficient to make a reliable
                      judgment (e.g., sparse events, conflicting signals)

\end{promptbox}

\begin{promptbox}{Active Problems Predictor}

=== INSTRUCTIONS ===

Identify active clinical problems or emerging risks this patient faces going
forward from this window.
- Only include risks that are real and imminent or already developing --- not
  distant or hypothetical
- Each risk must be tied to specific trajectory evidence
- An empty list is expected and acceptable when no urgent risks are present

\end{promptbox}

\begin{promptbox}{Action Recommendation Predictor}

=== INSTRUCTIONS ===

1. Recommend up to {int(top_k_actions)} distinct actions that are clinically
   actionable in the next {float(prediction_horizon_hours):g}-hour horizon.
2. Only recommend actions that are clearly justified by the available data.
3. It is totally acceptable to return fewer than {int(top_k_actions)}. If data
   is insufficient to justify a recommendation with at least low confidence,
   omit it.
4. Order actions from highest to lowest clinical priority (rank 1 = most urgent).
5. Prioritize interventions with the highest expected impact on short-term
   stability and outcome.
6. Ground every recommendation strictly in the provided context. Do not infer
   or invent missing data.

\end{promptbox}

\begin{promptbox}{Red Flag Actions Predictor}

=== INSTRUCTIONS ===

Using the full trajectory and current window, identify any actions that should
be strictly avoided for this specific patient going forward.
- Only flag actions that a reasonable clinician might consider but would be
  harmful for this specific patient --- do not list generic contraindications
  unless directly applicable here
- Each flag must be justified by patient-level evidence (comorbidities,
  trajectory events, organ function, known sensitivities)
- An empty list is expected and acceptable when no red flags are present

\end{promptbox}

\endgroup

\FloatBarrier